\documentclass[journal]{IEEEtran}

\usepackage{epstopdf}

\usepackage[cmex10]{amsmath}
\usepackage{mathtools} 
\usepackage{color}

\usepackage{array}

\usepackage{stfloats}
\usepackage{cite}
\usepackage{algorithm}
\usepackage{algpseudocode}
\usepackage{amsfonts}
\usepackage{tikz}
\usetikzlibrary{shapes,arrows,shadows}
\newtheorem{thm}{Theorem}
\newtheorem{defn}{Definition} 
\newtheorem{lemma}{Lemma}
\newtheorem{remark}{Remark}
\usepackage{hyperref}
\usepackage{stmaryrd}
\usepackage{url}
\usepackage{mathtools}
\newcommand{\mkv}{-\!\!\!\!\minuso\!\!\!\!-}
\allowdisplaybreaks

\algnewcommand\algorithmicinput{\textbf{Input:}}
\algnewcommand\Input{\item[\algorithmicinput]}
\algnewcommand\algorithmicoutput{\textbf{Output:}}
\algnewcommand\Output{\item[\algorithmicoutput]}

\begin{document}


\title{Compression-Based Regularization with an Application to Multi-Task Learning}

\author{Matias Vera,~\IEEEmembership{Student Member,~IEEE,}
        Leonardo Rey Vega,~\IEEEmembership{Member,~IEEE,}
        \\and Pablo Piantanida,~\IEEEmembership{Senior Member,~IEEE}\thanks{The work of M. Vera  was supported by a Peruilh Ph.D. grant from Facultad de Ingenier\'ia - Universidad de Buenos Aires. The work of L. Rey Vega was supported by grant  PIP11220150100578CO.}
\thanks{M. Vera is with the Facultad de Ingenier\'ia - Universidad de Buenos Aires, Argentina (e-mail: mvera@fi.uba.ar).}
\thanks{L. Rey Vega is with CSC-CONICET and the Facultad de Ingenier\'ia - Universidad de Buenos Aires, Argentina (e-mail: lrey@fi.uba.ar).}
\thanks{P. Piantanida is with CentraleSup\'elec - CNRS - Université Paris-Sud, France (e-mail: pablo.piantanida@centralesupelec.fr).}}\maketitle

\begin{abstract}
This paper investigates, from information theoretic grounds, a learning problem based on the principle that any regularity in a given dataset can be exploited to extract compact features from data, i.e., using fewer bits than needed to fully describe the data itself, in order to build meaningful representations of a relevant content (multiple labels). We begin by introducing the noisy lossy source coding paradigm with the log-loss fidelity criterion which provides the fundamental tradeoffs between the \emph{cross-entropy loss} (average risk) and the information rate of the features (model complexity). Our approach allows an information theoretic formulation of the \emph{multi-task learning} (MTL) problem which is a supervised learning framework in which the prediction models for several related tasks are learned jointly from common representations to achieve better generalization performance. Then, we present an iterative algorithm for computing the optimal tradeoffs and its global convergence is proven provided that some conditions hold. An important property of this algorithm  is that it provides a natural safeguard against overfitting, because it minimizes the average risk taking into account a penalization induced by the model complexity.  Remarkably, empirical results illustrate that there exists an optimal information rate minimizing the \emph{excess risk} which depends on the nature and the amount of available training data. An application to hierarchical text categorization is also investigated, extending previous works.
\end{abstract}


\IEEEpeerreviewmaketitle

\section{Introduction} \label{sec:intro}

The data deluge of the recent decades leads to new expectations for scientific discoveries from massive data in biology, particle physics, social media, safety and e-commerce. While mankind is drowning in data, a significant part of it is unstructured; hence it is difficult to discover relevant information. A common denominator in these novel scenarios is the challenge of representation learning: how to extract salient features or statistical relationships from data in order to build meaningful representations of the relevant content. 

Statistical models are used to acquire knowledge from data by identifying relationships between variables that allows making predictions and assessing their accuracy. The actual goal of learning is neither accurate estimation of model parameters nor compact representation of the data itself; rather, we are interested in the generalization capabilities, i.e., its ability to successfully apply rules extracted from previously seen data to characterize unseen data. It is known that complex models tend to produce \emph{overfitting}, i.e., represent the training data too accurately, therefore diminishing their ability to handle unseen data. To palliate this inconvenient, regularization methods include parameter penalization, noise, and averaging over multiple models trained with different sample sets. Nevertheless, it is not clear how to optimally control model complexity and therefore, this problem is an active research topic.

Shannon's seminal work~\cite{Shannon1993CodingTheoremsForADiscreteSourceWithAFidelityCriterion} on information compression with a fidelity criterion provides a function for measuring the distortion (or loss) between the original signal and its compressed representation. The rate-distortion function is related to a similarity measure in unsupervised learning/cluster analysis and has already demonstrated substantial performance improvement over standard supervised and unsupervised learning methods in a variety of important applications including compression, estimation, pattern recognition and classification, and statistical regression (see~\cite{726788} and references therein). This paper is concerned with an iterative algorithm for computing the rate-distortion of a generalization of Shannon's model, referred to as noisy source coding with the log-loss fidelity and side information, and its applications to multi-task learning.

\subsection{Related work}

The noisy source coding problem was first introduced by Dobrushin and Tsybakov~\cite{1057738} with the goal of generating a good description of an observed source $Y$ (at the encoder) in order to minimize its average distortion with respect to its reconstructed version (at the decoder). The main difference with respect to the original Shannon's problem relies on that $Y$ is not observed directly at the encoder. Instead, a noisy version of $Y$ denoted by $X$ is observed and appropriately compressed. More precisely, for a memoryless source with single-letter distribution $P_Y$ observed through a noisy channel with single-input transition probability $P_{X|Y}$, the noisy distortion-rate function under an additive distortion measure is given by
\begin{equation}
L(R,P_{XY}) \coloneqq  \min\limits_{P_{U|X}: \, I(U;X) \leq R }\mathbb{E}_{P_{XY}P_{U|X}}  [\ell(U,Y)]. \label{rate-distortion} 
\end{equation}
Motivated by the fact that it is not always obvious what loss function $\ell(u,y)$ should be used, especially if the data cannot be structured in a metric space (e.g. speech),  Tishby~\emph{et al.}~\cite{tishby99} associated this information-theoretic setup to a learning problem in which the encoder builds a (compressed) feature $U$ by extracting from data $X$ information about another variable $Y$. The idea of the so-called  \emph{Information Bottleneck} (IB) method is to identify relevant information from observed samples as being the information that those observations provide about another hidden signal (e.g., the information that face images $X$ provide about the names of the people portrayed $Y$). To this end, the IB introduces the \emph{log-loss fidelity}: 
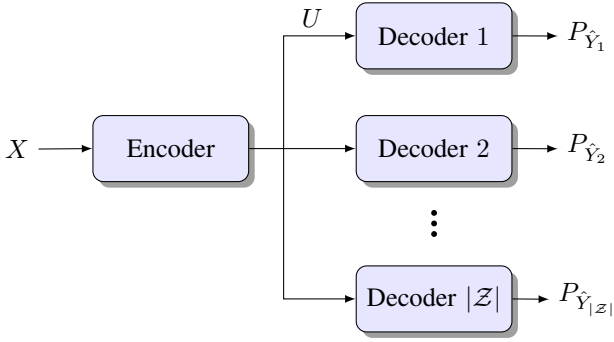
\begin{figure}[t!]
\begin{center}
   \pgfdeclarelayer{background}
   \pgfdeclarelayer{foreground}
   \pgfsetlayers{background,main,foreground}
   \tikzstyle{daemon}=[draw, fill=blue!20, text width=2.5em, text centered, drop shadow]
   \tikzstyle{dots} = [above, text centered]
   \tikzstyle{wa} = [daemon, fill=blue!10, minimum height=2.5em, rounded corners, drop shadow, inner sep=0.1em]
	\tikzstyle{arrow}=[draw, -latex] 
   \def\blockdist{1.3}
   \def\edgedist{1.5}
\begin{tikzpicture}
\node (xc) [wa,text width=2.0cm] {Encoder};
\path (xc)+(3.5,0.0) node (yc)[wa,text width=2.0cm] {Decoder $2$};
\path (xc)+(3.5,1.5) node (yc2)[wa,text width=2.0cm] {Decoder $1$};
\path (yc)+(0.0,-1.3) node (xi)[dots] {\huge $\vdots$};
\path (xc)+(3.5,-2.0) node (yc3)[wa,text width=2.0cm] {Decoder $|\mathcal{Z}|$};
\path (xc)+(1.5,0.0) node (naux1)[above] {};
\path (xc.west)+(-1.0,-0.25) node (xi)[dots] {$X$};
\path (yc.east)+(1.0,-0.32) node (yi)[dots] {$P_{\hat{Y}_2}$};
\path (yc2.east)+(1.0,-0.32) node (yi2)[dots] {$P_{\hat{Y}_1}$};
\path (yc3.east)+(1.0,-0.35) node (yi3)[dots] {$P_{\hat{Y}_{|\mathcal{Z}|}}$};
\path [arrow] (xi.east) -- node [above] {} (xc.west) ;
\path [arrow] (xc.east) -- node [above] {} (yc.west) ;
\path [arrow] (naux1.south) |- node [above,pos=0.7] {$U$} (yc2.west);
\path [arrow] (naux1.south) |- node [above] {} (yc3.west);
\path [arrow] (yc.east) -- node [above] {} (yi.west) ;
\path [arrow] (yc2.east) -- node [above] {} (yi2.west) ;
\path [arrow] (yc3.east) -- node [above] {} (yi3.west) ;
 \end{tikzpicture}
 \end{center}
\caption{Multi-task learning problem where $X$ denotes the input data, $U$ is the common feature (data representation) and  the corresponding soft-estimates of the multiple labels are denoted by $P_{\hat{Y}_1|U}, \dots, P_{\hat{Y}_{|\mathcal{Z}|}|U}$.}
\label{fig:db1}
\end{figure}  
\begin{equation}
\ell(u,y) \coloneqq   - \log  P_{Y|U} (y|u),\label{log-loss}
\end{equation}
where the \emph{soft-decoder} probability is obtained as:
\begin{equation}
P_{Y|U} (y|u)  =  \frac{\sum_{x\in\mathcal{X}}  P_{U|X} (u|x) P_{XY} (x,y)}{\sum_{x\in\mathcal{X}}
P_{U|X}(u|x)P_X(x)}, 
\label{soft-decoder}
\end{equation}
which is clearly determined by the \emph{soft-encoder} $P_{U|X}$ and the data distribution $P_{XY}$.  The optimal $P_{U|X}^\star$ is computed as 
the solution minimizing (\ref{rate-distortion}). This problem can be formulated using duality in optimization theory~\cite{boyd} which leads to:
\begin{equation}
 P_{U|X}^\star \coloneqq   \arg\min\limits_{P_{U|X}}  \mathbb{E}_{\hat{P}_{XY}P_{U|X}}[- \log  P_{Y|U} (Y|U) ] + \beta I(U;X), \label{tradeooff-classical}
\end{equation}
where the Lagrange multiplier $\beta\geq 0$ is a parameter that controls the tradeoff between compression rate $R$ and the average log-loss. However, the expectation is taken w.r.t. the sampling distribution $\hat{P}_{XY}$ because in real-world problems the true data distribution $P_{XY}$ is not known. Notice that the Lagrange multiplier $\beta$ can be interpreted as a parametrization of rate $R$. In this sense, it is clear that there is a family of optimal solutions, one for each $R$ or equivalently $\beta$.

An interesting variation is given by the observation that in the above problem the decoder is completely determined by the encoder and the data distribution. We can however consider that both $(P_{U|X},P_{\hat{Y}|U})$ have to be optimized:
\begin{eqnarray}
 (P_{U|X}^\star, P_{\hat{Y}|U}^{\star})&\!\coloneqq  \!&\arg\!\!\min\limits_{P_{U|X},P_{\hat{Y}|U}}\!\!\!  \mathbb{E}_{\hat{P}_{XY}P_{U|X}}[- \log  P_{\hat{Y}|U} (Y|U) ] \nonumber\\
& &+ \beta I(U;X), \label{tradeooff-classical2}
\end{eqnarray}
which  can be seen as the optimization of a penalized \emph{cross-entropy} metric\footnote{The cross-entropy  is a very common and popular cost function in machine learning (see~\cite{38136} and references therein)}. Different from (\ref{tradeooff-classical}), this problem does not lead (at least to our knowledge) to an information theory operational meaning as (\ref{tradeooff-classical}). However, it is easy to check that given an arbitrary  encoder $P_{U|X}$, the optimal decoder choice is given by (\ref{soft-decoder}). Therefore, expression~(\ref{tradeooff-classical2}) is --from the point of view of the optimization problem-- not more general than (\ref{tradeooff-classical}). For this reason and the connection with noisy source coding with log-loss fidelity, we will be concentrate our efforts in~(\ref{tradeooff-classical}).

Witsenhausen and Wyner \cite{witsenhausen75} were the first studying an information-theoretic  problem equivalent to~\eqref{rate-distortion} and obtained an interesting characterization of its solution and several applications to source coding. Whereas the IB method, in the same way as we presented it above, was introduced in~\cite{tishby99} as a rate-distortion problem with an additive fidelity measure.  Since then, it was applied to derive several clustering algorithms for a wide variety of applications such as: text classification~\cite{slonim2001}, galaxy spectra classification~\cite{slonim2001_2}, speaker recognition~\cite{hecht2009}, among others. Further information-theoretic extensions of the IB were recently considered in~\cite{our_isit15, tishby15, 2016arXiv160401433V, DBLP:journals/corr/PichlerPM16, DenizITW2017}. In particular, in some of these works it is shown that the same characterization of the rate and distortion tradeoff is obtained when the distortion metric is not necessarily additive as assumed in~\eqref{log-loss}. 
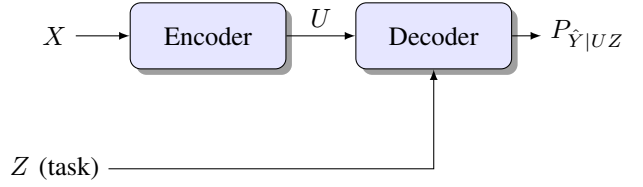
\begin{figure}[t!]
\begin{center}
   \pgfdeclarelayer{background}
   \pgfdeclarelayer{foreground}
   \pgfsetlayers{background,main,foreground}
   \tikzstyle{daemon}=[draw, fill=blue!20, text width=2.5em, text centered, drop shadow]
   \tikzstyle{dots} = [above, text centered]
   \tikzstyle{wa} = [daemon, fill=blue!10, minimum height=2.5em, rounded corners, drop shadow, inner sep=0.1em]
	\tikzstyle{arrow}=[draw, -latex] 
   \def\blockdist{1.3}
   \def\edgedist{1.5}
\begin{tikzpicture}
\node (xc) [wa,text width=2.0cm] {Encoder};
\path (xc)+(3.0,0.0) node (yc)[wa,text width=2.0cm] {Decoder};
\path (xc.west)+(-1.0,-0.244) node (xi)[dots] {$X$};
\path (yc.east)+(1.0,-0.33) node (yi)[dots] {$P_{\hat{Y}|UZ}$};
\path (xi.south)+(0.0,-1.8) node (zi)[dots] {$Z$ (task)};
\path [arrow] (xi.east) -- node [above] {} (xc.west) ;
\path [arrow] (xc.east) -- node [above] {$U$} (yc.west) ;
\path [arrow] (zi.east) -| node [above] {} (yc.south);
\path [arrow] (yc.east) -- node [above] {} (yi.west) ;
 \end{tikzpicture}
 \end{center}
\caption{Multi-task learning as the noisy source coding problem with log-loss fidelity and side information $Z$ (index task variable) at the decoder, referred to as the IB with side information. }
\label{fig:db2}
\end{figure}\par
The optimization in~\eqref{tradeooff-classical} was typically addressed by resorting to Blahut--Arimoto (BA) type algorithms. These are often used to refer to a class of algorithms for numerically computing the capacity of a noisy channel and the rate-distortion function for given source. They are iterative algorithms that eventually converge to the optimal solution provided that the  optimization problem is convex. The algorithm for the classical rate-distortion problem, i.e., by setting $X=Y$ in~\eqref{rate-distortion}, was developed independently by Arimoto~\cite{arimoto72} and Blahut~\cite{blahut72}.  An extension of this algorithm to the rate-distortion function with side information at the decoder was reported in~\cite{willems83}. Although this algorithm can be applied for optimizing the IB criterion~\cite{tishby99}, we emphasize that conventional algorithms~\cite{arimoto72,blahut72} are only expected to converge to a local minimum since expression~\eqref{tradeooff-classical} leads to a non-convex problem due to the presence of the soft-encoder in the fidelity measure in~\eqref{log-loss}. Chechik \emph{et al.}~\cite{chechik02} adapts a BA algorithm to a restricted form of side information without further study the involved algorithm. In a different but related optimization problem, Kumar and Thangaraj~\cite{kumar08} adapt the BA algorithm and analysis techniques provided in~\cite{yasui07} to a non-convex problem while Yasui and Matsushima~\cite{yasui10} extend this work for computing rate regions.\par
In this paper, we present a novel algorithm for multi-task learning  based on the IB paradigm with side information. Multi-task Learning (MTL)~\cite{Caruana1997} is an approach to inductive transfer that improves generalization by using the domain information contained in the training signals of related tasks as an inductive bias. This is accomplished by learning tasks in parallel while using a shared data representation, as described in Fig.~\ref{fig:db1}. What is learned for each task can help other tasks to be learned better and thus can result in improved efficiency and prediction accuracy when compared to training the models separately~\cite{Baxter00}. MTL has received a great deal of attention in the recent years \cite{zhang_survey_2017}. There are basically two ways of improving generalization via MTL. One approach imposes a structural condition  on the learned  parameters for all related tasks, e.g., by assuming some low-rank structure~\cite{ando_framework_2005} or by modelling explicitly the  links between tasks~\cite{ciliberto_convex_2015}. The other approach is through learning of common features for all desired tasks~\cite{argyriou_convex_2008} via a common encoder (or feature selector) followed by a task-specific predictor, e.g., using a different decoder for each task. The later is the one we investigate in this paper. However, our setup differs from previous works in that we focus an information-theoretic formulation of the MTL problem. We should also mention that we restrict our setup to MTL scenarios  where the inputs are common to all tasks. Although this can be mathematically equivalent to the problem of multi-label learning (MLL), there are some important differences (see~\cite{zhang_review_2014} for further details). 

\subsection{Our contribution}

We first introduce an information-theoretic paradigm which provides the fundamental tradeoff between the \emph{log-loss} (average risk) and the information rate of the features (statistical model complexity). We derive an iterative Arimoto-Blahut like algorithm to address the non-convex optimization problem of the IB method in presence of side information available only at the decoder~\cite{wyner_ziv76, our_isit15}, as described in Fig.~\ref{fig:db2}, and prove its global convergence provided that some conditions hold. It worth to mention that our formulation, as a noisy source coding problem with side information at the decoder, provides an information-theoretic perspective to the MTL problem, which yields a valuable connection between the fields of machine learning and Shannon theory. In precise terms, the encoder aims at extracting relevant (common) information $U$ from a data set $X$ about labels $Y|Z=z$ needed for a collection of tasks $z\in\mathcal{Z}$ at the decoder. The function cost weights of each of these tasks will be defined to be the probability mass function of a randomly chosen index task variable $Z$. The representation $U$ is expected to summarize data $X$ in a compact way, where compactness of the model is measured in terms of the minimum Shannon entropy rate. However, learning a representation $U$ for predicting $Y$ requires to capture the regularities in $Y$ that are present in $X$ while other irrelevant information for $Y$ must be disregarded. In this sense, our statistical measure of complexity says that the best description $U$ of the data is given by the model that compresses $Y$ the best which is captured by Shannon mutual-information rate $I(U;X|Z=z)$. In the spirit of the Kolmogorov-Chaitin complexity~\cite{Li97anintroduction} that is a measure of the regularities present in an object above and beyond pure randomness. This approach provides a natural safeguard against overfitting by minimizing an average risk penalized by the model complexity.  Remarkably, empirical results illustrate that there exists an optimal information rate minimizing the \emph{excess risk} which depends on the nature and the amount of available training data. We further evaluate the performance of this algorithm on hierarchical text categorization of documents  and numerical results demonstrates the merits of the proposed MTL algorithm in terms of the classification performance.

The rest of the paper is organized as follows. In Section~\ref{sec:problem}, we introduce the problem and present our iterative algorithm. The algorithm's properties are analyzed in Section~\ref{sec:algoritmo} while in Section~\ref{sec:experimentos} we show numerical evidence for some selected applications. Section~\ref{sec:conclusiones} provides concluding remarks and major mathematical details are relegated to Appendices.

\section{Problem Definition and Main Result}
\label{sec:problem}

\subsection{Notation and conventions}
We use upper-case letters to denote random variables and lower-case letters to denote realizations of random variables (RVs). Superscripts are used to denote the length of the vectors and subscripts denote the index of the components of a vector. The probability mass function (pmf) of random variable $X$ is denoted by $P_X(x)$, $x\in\mathcal{X}$, where $\mathcal{X}$ is the  alphabet of the random variable. When clear from the context we will simply refer to the pmf of $X$ as $P_X$. All alphabets are assumed finite. $\mathbb{E}_{P_X}[\cdot]$ denotes the expectation and $|\mathcal{A}|$ indicates the cardinality of a set $\mathcal{A}$. $A\mkv B\mkv C$  indicates a Markov chain, i.e., ${P}_{A|BC}= {P}_{A|B}$. The support of a pmf $P_X$ is denoted by $\text{supp}(P_X)$. The information measures to be used are~\cite{cover}: the \emph{entropy}  $H(X)\coloneqq   \mathbb{E}_{P_X}\left[-\log { {P}_X}(X)\right]$,  the \emph{conditional  entropy} $H(X|Y)\coloneqq   \mathbb{E}_{P_{XY}}\left[-\log {P}_{X|Y} (X|Y)\right]$ and the \emph{relative entropy}: 
\begin{equation}
\mathcal{D}( {P}_{X}\| {Q}_{X})=\left\{  
\begin{array}{ll}
\displaystyle \mathbb{E}_{P_X}\left[\log \frac{ {P}_{X}(X)}{ {Q}_X(X)}\right] &  \textrm{if $P_X \ll Q_X$}\\
+\infty &  \textrm{otherwise,}
\end{array}
\right.
\end{equation}
where we use $P_X \ll Q_X$ to denote that the probability measure $P_X$ is \emph{absolutely continuous} w.r.t. $Q_X$, and the \emph{mutual information}: $I(X;Y)\coloneqq  \mathcal{D}(  {P}_{XY} \| { {P}_X {P}_Y})$. When referring to an empirical distribution computed using data samples we will use notation $\hat{P}_X$. Functionals computed with an empirical distribution will be also denoted similarly, e.g., the entropy of $X$ computed by using $\hat{P}_X$ is denoted as: $\hat{H}(X)$. All logarithms are assumed to be base $2$.

\subsection{Multi-task learning and Information Bottleneck}

Let $(X,Y,Z)$ be RVs with joint probability mass function $P_{XYZ}$. A soft-encoder $P_{U|X}$  wishes to extract from $X$ information about a collection of labels $Y|Z=z$ with $z\in\mathcal{Z}$ while the randomly chosen task with index $Z$ is available only at the decoder, as shown in Fig.~\ref{fig:db2}. 

Following our previous discussions right after (\ref{tradeooff-classical2}), the optimal soft-decoder $P_{Y|UZ}$ will depend on the selected  $(P_{U|X},P_{XYZ})$  and is given by 
\begin{equation}\label{ec:decoder}
P_{Y|UZ}=\frac{\sum_xP_{U|X}P_{XYZ}}{\sum_xP_{U|X}P_{XZ}}.
\end{equation}
We focus on the average \emph{log-loss risk} that coincides with the conditional entropy: 
\begin{equation}
\mathbb{E}_{P_{XYZ}P_{U|X} }[- \log  P_{Y|UZ} (Y|UZ)] = H(Y|UZ). 
\end{equation}
Finding the encoder that minimizes the average log-loss is equivalent to search for the encoder maximizing the mutual (relevance) information $I (Y;U|Z)=H(Y|Z)-H(Y|UZ)$. As a consequence, we can focus on maximizing the relevance (mutual information) $I (Y;U|Z)$ subject to a given complexity (Shannon rate) $I (X;U|Z)$. As a matter of fact, it has been shown in~\cite[Theorem 1 with $\mu_2=0$, $L=1$]{our_isit15}  that this tradeoff corresponds  to the best possible asymptotically (over the block-length) achievable tradeoff between the multi-letter relevance and the compression rate.  

\begin{defn}[Relevance-rate region]\label{def:singleletter}
A pair rates $(R,\mu)$ is achievable iff it belongs to the rate-relevance region:
\begin{align}
\label{laregion}
\mathcal{R}\coloneqq  \big\{(\mu,R)&\in\mathbb{R}^2_{\geq0}:\exists\;P_{U|X}\textrm{ s.t. }\ R \geq I(X;U|Z),\nonumber\\
\mu& \leq I(Y;U|Z),\quad U\mkv X\mkv(Y,Z)\big\},
\end{align}
and the corresponding relevance-rate function is defined by
\begin{eqnarray}
L(R,P_{XYZ}) &=& \max \big\{\mu:\, \, (R,\mu)\in \mathcal{R} \big\},\\
&=& \max\limits_{P_{U|X}: \, I(U;X|Z) \leq R } I(U;Y|Z).
\label{eq:rel_rate_funct}
\end{eqnarray}
\end{defn}
\begin{figure}[t!]
	\begin{center}
		\pgfdeclarelayer{background}
		\pgfdeclarelayer{foreground}
		\pgfsetlayers{background,main,foreground}
		\def\blockdist{1.3}
		\def\edgedist{1.5}
		\includegraphics[width=0.45\textwidth]{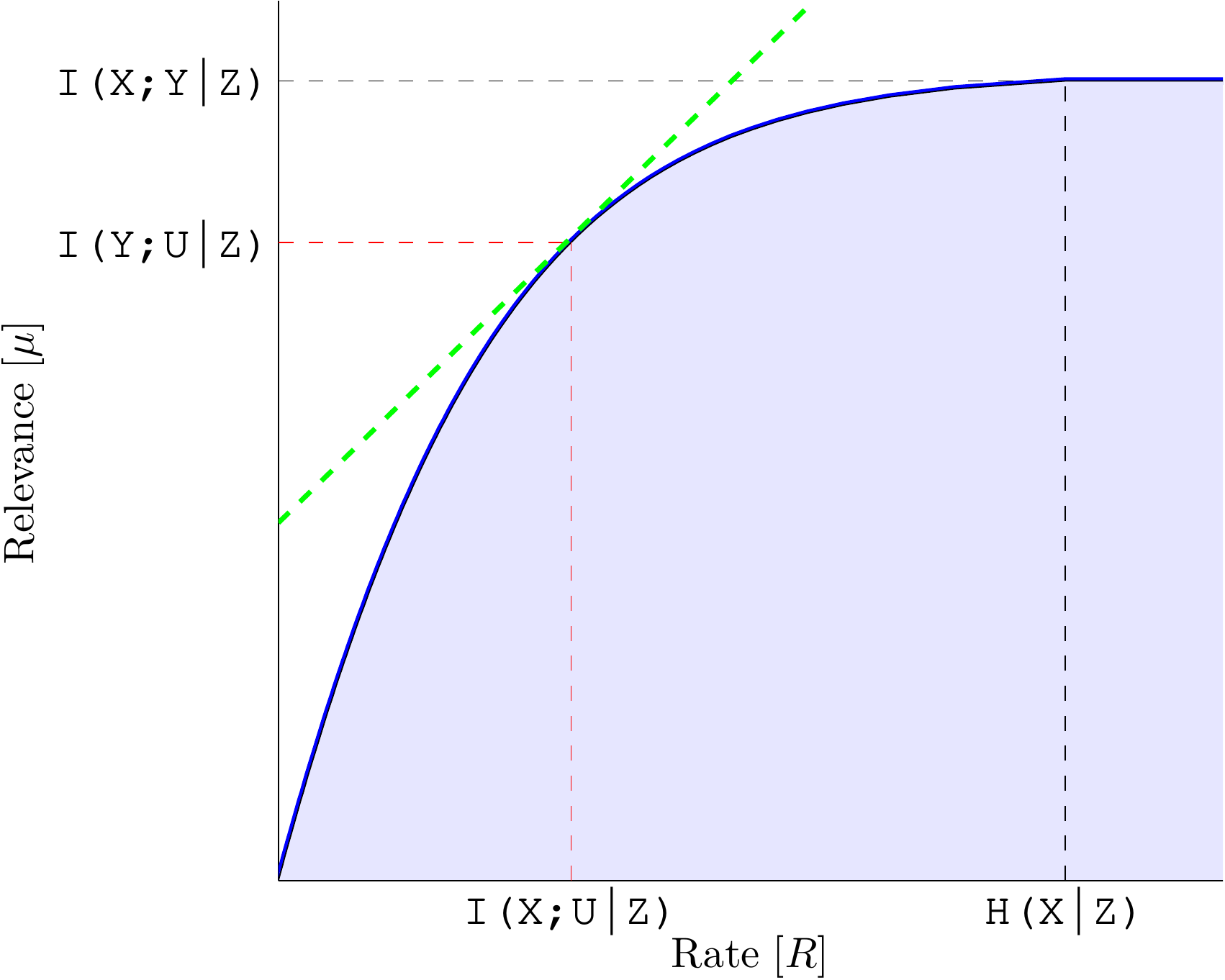}
		\caption{A relevance-rate region with its supporting hyperplane~\eqref{problema}.}
		\label{fig:region}
		\begin{tikzpicture}[remember picture,overlay]
		\path (-2.6,3.7) node (ast) [above] {$\frac{V_\lambda}{\lambda}$};
		\draw [red,line width=0.3mm,fill=red]  (0.85,7.05) node[anchor=north]{}
		-- (1.15,7.05) node[anchor=north]{}
		-- (1.15,7.35) node[anchor=south]{}
		-- cycle;
		\path (1.7,7.5) node (ast) [above] {$\frac{1-\lambda}{\lambda}$};
		\draw [red, -latex', line width=1pt] (1.1,7.2) to[out=0,in=270,distance=0.4cm] (1.6,7.5);
		\end{tikzpicture}
	\end{center}
\end{figure}
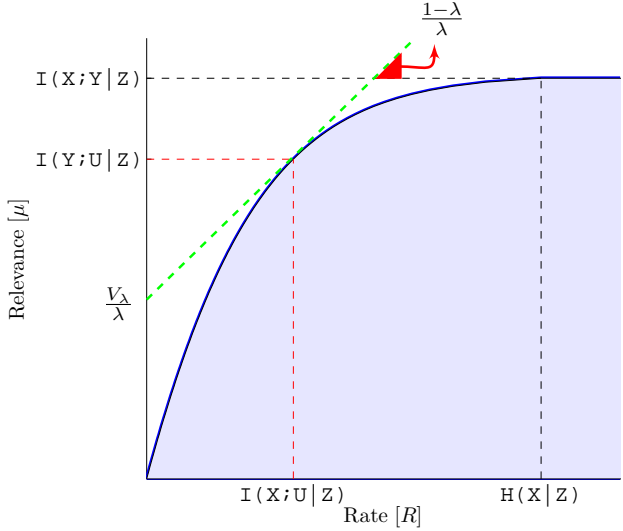
\begin{lemma}
\label{lem:cvx}
$\mathcal{R}$ is closed, convex and the cardinality of random variable $U$ can be bounded as $|\mathcal{U}|\leq|\mathcal{X}|+1$ without loss of generality.
\end{lemma}
\begin{IEEEproof}
See Appendix \hyperref[app_cvx]{A}.
\end{IEEEproof}

Observe that the relevance-rate function --as being the upper-boundary of $\mathcal{R}$-- provides an alternative and complete characterization of the region. It is important to mention that the maximum in this problem is well-defined because we are attempting to maximize a continuous function over a compact set. An example of the relevance-rate region can be seen in Fig. \ref{fig:region}. The relevance-rate function quantifies the maximum mutual information (conditioned on $Z$) between $Y$ and the generated description $U$ (using solely $X$) when a bound on the description complexity (rate) is imposed on $U$. 

Although the optimization involved  in~\eqref{eq:rel_rate_funct} does not lead to a convex problem, the properties of $\mathcal{R}$ allows us to characterize the optimal tradeoff between compression and relevance rates using \emph{supporting hyperplanes}~\cite{boyd}. As it is well known, any closed and convex set can be characterized from all its supporting hyperplanes~\cite{rockafellar_convex_1970}. A supporting hyperplane for $\mathcal{R}$ with parameter $\lambda$ can be written as:
\begin{equation}
V_\lambda\coloneqq  \max_{P_{U|X}}\;\lambda I(Y;U|Z)-(1-\lambda)I(X;U|Z).
\label{problema}
\end{equation}
With little effort it is easy to show that $\lambda\in[0,1]$ suffices for the full characterization of $\mathcal{R}$ using supporting hyperplanes. 
\begin{figure*}[th!]
	\footnotesize
	\hrulefill\\
	{\bf ALGORITHM 1:} Information Bottleneck with side information.\par
	\hrulefill
	\begin{algorithmic}[*]
		\Input $P_{XYZ}$, $P_{U|X}^{(0)}$, $\lambda\in[0,1]$, $\epsilon>0$.
		\Output $P_{U|X}^{*,\lambda}$.
		\State Initialize $n \coloneqq  0$, $I_\lambda^{(0)} \coloneqq  +\infty$, $F_{\lambda}^{(0)} \coloneqq  0$.
		\While {$I_\lambda^{(n)}-F_{\lambda}^{(n)}>\epsilon$}
		\vspace{2mm}
		\State Compute 
		$$Q_{U|YZ}^{(n+1)} \coloneqq  \sum_xP_{U|X}^{(n)}P_{X|YZ},\;\;\;Q_{X|ZU}^{(n+1)} \coloneqq  \frac{P_{U|X}^{(n)}P_{X|Z}}{\sum_{x^\prime}P_{U|X^\prime}^{(n)}P_{X^\prime|Z}},$$
		$$P_{U|X}^{(n+1)} \coloneqq  k_x\cdot \exp\left\{\frac{2\lambda-1}{\lambda}\sum_{z}P_{Z|X}\log( Q_{X|ZU}^{(n+1)})+\sum_{y,z}P_{Y|XZ}P_{Z|X}\log( Q_{U|YZ}^{(n+1)})\right\},$$
		$$F_\lambda^{(n+1)} \coloneqq  (2\lambda-1)\sum_{x,z,u}{P}_{U|X}^{(n+1)}{P}_{XZ}\log\left(\frac{{Q}_{X|ZU}^{(n+1)}}{{P}_{X|Z}}\right)-\lambda\sum_{x,y,z,u}{P}_{U|X}^{(n+1)}{P}_{XYZ}\log\left(\frac{{P}_{U|X}^{(n+1)}}{{Q}_{U|YZ}^{(n+1)}}\right),$$
		$$I_\lambda^{(n+1)} \coloneqq  \max_{u}\;(2\lambda-1)\sum_{x,z}{P}_{XZ}\log\left(\frac{{Q}_{X|ZU}^{(n+1)}}{{P}_{X|Z}}\right)-\lambda\sum_{x,y,z}{P}_{XYZ}\log\left(\frac{{P}_{U|X}^{(n)}}{{Q}_{U|YZ}^{(n+1)}}\right),$$
		\State Update $n \coloneqq  n+1$.
		\EndWhile \vspace{0.2cm}
		\State Report $P_{U|X}^{*,\lambda}=P_{U|X}^{(n)}$.\\
		\vspace{0.2cm}
		\hrulefill
	\end{algorithmic}
\end{figure*}

Finding the optimal encoder $P_{U|X}^{*,\lambda}$ in \eqref{problema} requires knowledge of the underlying distribution $P_{XYZ}$. In practical applications, this lack of knowledge is overcome by resorting to labeled examples, i.e., a training set of $n$ i.i.d. tuples: $\{(x_1,y_1,z_1),\dots, (x_n,y_n,z_n)\}$ sampled according to the unknown distribution $P_{XYZ}$. In Section~\ref{sec:experimentos},  we will study some supervised learning setups where  expression~\eqref{problema} together with the iterative algorithm described below will serve as a supervised objective to guide multi-task learning. 

\subsection{An iterative optimization algorithm}

In order to simplify the notation, we define $f(\lambda,P_{U|X})$ as:
\begin{equation}
f(\lambda,P_{U|X})\coloneqq  \lambda I(Y;U|Z)-(1-\lambda)I(X;U|Z).
\end{equation}
Clearly, we can write
\begin{align}
f(\lambda,P_{U|X})=&\sum_{z\in\mathcal{Z}}P_Z(z)\big[\lambda I(Y;U|Z=z)\nonumber\\
&-(1-\lambda)I(X;U|Z=z)\big],
\end{align}
where we see the effect of the weights $P_Z(z)$ associated with each task. \emph{Data Processing Inequality} \cite[sec.~2.3]{gamal} allows to conclude that the only allowable values are $0\leq\mu\leq I(X;Y|Z)$. We wish to obtain an algorithm that is able to find the supporting hyperplanes of $\mathcal{R}$, for every $\lambda\in[0,1]$, allowing the computation of the upper-boundary of $\mathcal{R}$, i.e., finding the optimal pmf $P_{U|X}^{*,\lambda}$ that achieves the maximum in~\eqref{problema}, and evaluating the corresponding mutual informations $I^{*,\lambda}(X;U|Z)$ and $I^{*,\lambda}(Y;U|Z)$ based on:
\begin{equation}
P_{U|X}^{*,\lambda}\coloneqq   \arg\max_{P_{U|X}}\;f(\lambda,P_{U|X}).
\end{equation}
By relying on the Markov chain $U\mkv(X,Z)\mkv Y$ implied from $U\mkv X\mkv(Y,Z)$, the function $f(\lambda,P_{U|X})$ writes as:
\begin{equation}
\label{problemv2}
f(\lambda,P_{U|X})=(2\lambda-1) I(X;U|Z)-\lambda I(X;U|Y,Z).
\end{equation}
Depending on the value of $\lambda$, it is appropriate to define the algorithm in two different ways. This is similar to the approach in \cite{yasui10}. If $\lambda\in[0,0.5]$, both terms of \eqref{problemv2} are non-positive and thus, the solution is trivial: $V_\lambda=0$. This is achieved for all pmf that satisfies $P_{U|X}=P_{U}$ and corresponds to the point $(0,0)$ in $\mathcal{R}$. The relevant case is when $\lambda\in(0.5,1]$. In this case, the proposed iterative  algorithm is summarized in Alg.~\hyperref[algo]{1}, where $k_x$ are constants such that $\sum_uP_{U|X}^{(n+1)}=1\;\forall x\in\mathcal{X}$. In the next section, we explain the rationale behind this algorithm.

\section{Algorithm Analysis}
\label{sec:algoritmo}

The problem of finding the global maximum  is not convex because $P_{U|X} \mapsto f(\lambda, P_{U|X})$ is not concave. As a consequence, we cannot expect to have an efficient procedure that allow us to finding the global maximum of the problem. The algorithm proposed is a variant of the BA algorithm~ \cite{blahut72,arimoto72} which is based on solid theoretical grounds and guarantee global optimum convergence results when the optimization problem is convex (e.g. the capacity and rate-distortion functions). Although our optimization problem is not convex and thus, the general convergence to the global optimum cannot be guaranteed, we derive results regarding the convergence to the global maximum provided that some additional conditions are fulfilled. Our results are inspired from seminal works in~\cite{kumar08,yasui07,yasui10}. A convergence rate result is also derived in Appendix~\hyperref[app:convergence-rate]{D}. 

\subsection{Algorithm summary}
We first study the algorithms expressions in further detail. Eq. \eqref{problemv2} can be expanded as:
\begin{align}
f(\lambda,P_{U|X})&=(2\lambda-1)\sum_{x,z,u}P_{U|X}P_{XZ}\log\left(\frac{P_{X|ZU}}{P_{X|Z}}\right)\nonumber\\
&\quad-\lambda\sum_{x,y,z,u}P_{U|X}P_{XYZ}\log\left(\frac{P_{U|X}}{P_{U|YZ}}\right).
\end{align}
Let the function ${F}(\lambda,P_{U|X}, Q_{U|YZ}, Q_{X|ZU})$ be:
\begin{align}
&{F}(\lambda,P_{U|X}, Q_{U|YZ}, Q_{X|ZU})\nonumber\\
&\qquad\coloneqq  (2\lambda-1)\sum_{x,z,u}P_{U|X}P_{XZ}\log\left(\frac{ Q_{X|ZU}}{P_{X|Z}}\right)\nonumber\\
&\qquad\quad-\lambda\sum_{x,y,z,u}P_{U|X}P_{XYZ}\log\left(\frac{P_{U|X}}{ Q_{U|YZ}}\right),
\end{align}
where $Q_{U|YZ}, Q_{X|ZU}$ are arbitrary pmfs. For sake of simplicity, sometimes we write ${F}$ when the arguments are obvious. This new function has some important properties.

\begin{lemma}
\label{fprop}
Consider any $P_{U|X}$ and let $\lambda\in(0.5,1]$. The following properties hold  true:
\begin{enumerate}
\item $f(\lambda,P_{U|X})\geq{F}(\lambda,P_{U|X}, Q_{U|YZ}, Q_{X|ZU})$, 
and equality is achieved iff $Q_{U|YZ}=P_{U|YZ}\;\forall\ (y,z)\in\mathcal{Y}\times\mathcal{Z}$ and $ Q_{X|ZU}=P_{X|ZU}\;\forall (z,u)\in\mathcal{Z}\times\mathcal{U}$.
\item The value $V_\lambda$ satisfies:
\begin{equation}
V_\lambda=\max_{P_{U|X}}\max_{ Q_{U|YZ}, Q_{Y|ZU}}{F}(\lambda,P_{U|X}, Q_{U|YZ}, Q_{X|ZU}).
\end{equation}
\item For any $Q_{U|YZ}, Q_{X|ZU}$ and $\lambda\in(0.5,1]$, $P_{U|X}\mapsto {F}(\lambda,P_{U|X}, Q_{U|YZ}, Q_{X|ZU})$ is concave and achieves its maximum provided that:
\begin{align}
P_{U|X}&= k_x\cdot \exp\Big\{\frac{2\lambda-1}{\lambda}\sum_{z\in\mathcal{Z}}P_{Z|X}\log( Q_{X|ZU})\nonumber\\
&\quad+\sum_{(y,z)\in\mathcal{Y}\times\mathcal{Z}}P_{Y|XZ}P_{Z|X}\log( Q_{U|YZ})\Big\},\label{eq:Pu_x}
\end{align} 
where $k_x$ are constants such that $\sum_uP_{U|X}=1\;\forall x\in\mathcal{X}$.
\end{enumerate}
\end{lemma}

\begin{IEEEproof}
	\begin{enumerate}
		\item The difference between functions  can be written as:
		\begin{align}
\hspace{-0.2cm}f&(\lambda,P_{U|X})-{F}(\lambda,P_{U|X}, Q_{U|YZ}, Q_{X|ZU})\nonumber\\
		&=(2\lambda-1)\sum_{x,y,z,u}P_{U|X}P_{XYZ}\log\left(\frac{P_{X|ZU}}{P_{X|Z}}\right)\nonumber\\
		&\quad-\lambda\sum_{x,y,z,u}P_{U|X}P_{XYZ}\log\left(\frac{P_{U|X}}{P_{U|YZ}}\right)\nonumber\\
		&\quad-(2\lambda-1)\sum_{x,y,z,u}P_{U|X}P_{XYZ}\log\left(\frac{ Q_{X|ZU}}{P_{X|Z}}\right)\nonumber\\
		&\quad+\lambda\sum_{x,y,z,u}P_{U|X}P_{XYZ}\log\left(\frac{P_{U|X}}{ Q_{U|YZ}}\right)\\
		&=\lambda\sum_{y,z}P_{YZ}\;\mathcal{D}(P_{U|Y,Z}\| Q_{U|YZ})\nonumber\\
		&\quad+(2\lambda-1)\sum_{z,u}P_{ZU}\;\mathcal{D}(P_{X|ZU}\| Q_{X|ZU})\geq 0
		\end{align}
with equality iff $Q_{U|YZ}=P_{U|YZ}\;\forall\;(y,z)\in\mathcal{Y}\times\mathcal{Z}$ and $ Q_{X|ZU}=P_{X|ZU}\;\forall\;(z,u)\in\mathcal{Z}\times\mathcal{U}$. This is easily seen from the properties of relative entropy~\cite[sec.~2.3]{gamal}.
		\item The claim follows by combining the previous claim with~\eqref{problema}.
		\item Every pmf satisfies $\sum_u P_{U|X}=1$. Then, using Lagrange multipliers $c_x,\ x\in\mathcal{X}$:
\begin{align}
&\frac{\partial[{F}+\sum_xc_x(\sum_uP_{U|X}-1)]}{\partial P_{U|X}}=c_x-\lambda P_X\log(eP_{U|X})\nonumber\\
&\quad+(2\lambda-1)\sum_{z}P_{XZ}\log\left( Q_{X|ZU}\right)\nonumber\\
&\quad+\lambda\sum_{y,z}P_{XYZ}\log( Q_{U|YZ})=0
\end{align}
from which we immediately recover \eqref{eq:Pu_x}. Note that this solution meet $P_{U|X=x}(u)\geq 0$ for all $(x,u)\in\mathcal{X}\times\mathcal{U}$. The concavity results follow from:
\begin{equation}
\hspace{-0.6cm}\frac{\partial^2{F}(\lambda,P_{U|X}, Q_{U|YZ}, Q_{Y|ZU})}{\partial P_{U|X}^2}=-\frac{\lambda P_{X}\log(e)}{P_{U|X}}\leq0.
\end{equation}
\end{enumerate}
\end{IEEEproof}

We observe that the function ${F}(\lambda,P_{U|X}, Q_{U|YZ}, Q_{X|ZU})$ provides an achievable and easy way to optimize a lower bound to the objective function $f(\lambda,P_{U|X})$, for each $P_{U|X}$. Interestingly, $P_{U|X}\mapsto {F}(\lambda,P_{U|X}, Q_{U|YZ}, Q_{X|ZU})$ is concave for each $(Q_{U|YZ}, Q_{X|ZU})$, guaranteeing that any local optimum is also a global one. These facts lead naturally to the iterative process in order to perform the double maximization which results in $V_\lambda$. This is the case in Algorithm 1, where we perform an iterative maximization process on both arguments: $P_{U|X}$ and $(Q_{U|YZ}, Q_{X|ZU})$. For a given $\lambda\in(0.5,1]$, starting from an initial condition $P_{U|X}^{(0)}$, and according to 2) in Lemma~\ref{fprop}, we  search for $Q_{U|YZ}^{(1)},Q^{(1)}_{X|ZU}$ such that the maximum of ${F}(\lambda,P_{U|X}^{(0)}, Q_{U|YZ}, Q_{X|ZU})$ is achieved, for fixed $P_{U|X}^{(0)}$. Next, from 3) in the previous lemma, we find ${P}_{U|X}^{(1)}$ as the argument that maximizes ${F}(\lambda,P_{U|X}, Q^{(1)}_{U|YZ}, Q^{(1)}_{X|ZU})$. This iterative process is repeated until a stopping criterion is satisfied (see Section~\ref{subsec:stopping}). It is easy to show that the sequence of values ${F}(\lambda,P_{U|X}^{(n)}, Q^{(n)}_{U|YZ}, Q^{(n)}_{X|ZU})$ is monotone non-decreasing. This clearly guarantees the convergence. In the sequel, we further study this process  in detail.

\subsection{Convergence analysis}

For sake of  simplicity, let us assume  that the optimal point $P_{U|X}^{*,\lambda}$ is unique. Define $F_\lambda^{(n)}\coloneqq {F}(\lambda,{P}_{U|X}^{(n)},{Q}_{U|YZ}^{(n)},{Q}_{X|ZU}^{(n)})$. From the previous section we know that $F_\lambda^{(n+1)}\geq F_\lambda^{(n)}$. Moreover, from 2) in Lemma~\ref{fprop}, $V_\lambda\geq F_\lambda^{(n)}$ for all $n$. However, there is no guarantee that $V_\lambda=F_\lambda^{(\infty)}$. In order to obtain some insights on the convergence process and on the limiting point of the iterative process, we will consider the concept of $\delta$-superlevel set (see~\cite{kumar08} for further details).

\begin{defn}
	The $\delta$-superlevel set is defined as the set:
	\begin{equation}
	G_{\delta,\lambda}\coloneqq \left\{{P}_{U|X}: \mathcal{X} \rightarrow \mathcal{P}(\mathcal{U})\big| \; f(\lambda,{P}_{U|X})\geq \delta\right\}.
	\end{equation}
\end{defn}

\begin{defn}
	Consider a fixed conditional distribution $\tilde{P}\in G_{\delta,\lambda}$. The set $H_{\delta,\lambda}(\tilde{P})$ is defined as the set of all points ${P}_{U|X}\in G_{\delta,\lambda}$ such that each of them (and $\tilde{P}$) are in the same path-connected component of $G_{\delta,\lambda}$. In order words, $H_{\delta,\lambda}(\tilde{P})$ is the set of all points ${P}_{U|X}\in G_{\delta,\lambda}$ that are reachable from $\tilde{P}$ by a continuous path.
	\label{def3}
\end{defn}

\begin{lemma}
	\label{adentro!}
Let $\lambda\in(0.5,1]$, the distribution ${P}_{U|X}^{(n+1)}$ lies in $H_{\delta,\lambda}({P}_{U|X}^{(n)})$ for all $k$ such that ${P}_{U|X}^{(n)}\in G_{\delta,\lambda}$.
\end{lemma}
\begin{IEEEproof}
Let $\tilde{G}_{\delta,\lambda}^n$ be the $k$-superlevel of the function ${F}(\lambda,{P}_{U|X},{Q}_{U|YZ}^{(n+1)},{Q}_{X|ZU}^{(n+1)})$. Since $f(\lambda,{P}_{U|X})\geq{F}(\lambda,{P}_{U|X},{Q}_{U|YZ}^{(n+1)},{Q}_{X|ZU}^{(n+1)})$ from Lemma \ref{fprop} (i.e. by claim 1), it follows that $\tilde{G}_{\delta,\lambda}^n\subseteq G_{\delta,\lambda}\ \forall n$. Also, ${P}_{U|X}^{(n)}$ and ${P}_{U|X}^{(n+1)}$ lies in $\tilde{G}_{\delta,\lambda}^n$ because:
\begin{equation}
F_\lambda^{(n+1)}\geq{F}(\lambda,{P}_{U|X}^{(n)},{Q}_{U|YZ}^{(n+1)},{Q}_{X|ZU}^{(n+1)})=f(\lambda,{P}_{U|X}^{(n)}).
\end{equation}
For fixed $({Q}_{U|YZ}^{(n+1)},{Q}_{X|ZU}^{(n+1)})$  pmfs, we know that ${F}$ is concave in argument ${P}_{U|X}$. Thus, $\tilde{G}_{\delta,\lambda}^n$ is a convex set and it is therefore path-connected and between any two of its points there exists a continuous path. Then, it follows that $\tilde{G}_{\delta,\lambda}^n\subseteq H_{\delta,\lambda}({P}_{U|X}^{(n)})$ and we conclude that  ${P}_{U|X}^{(n+1)}\in H_{\delta,\lambda}({P}_{U|X}^{(n)})$.
\end{IEEEproof}

Clearly, this lemma and Definition \ref{def3} imply that if ${P}_{U|X}^{(0)}\in G_{\delta, \lambda}$ for a given value of $\delta$, then ${P}_{U|X}^{(n)}\in H_{\delta, \lambda}({P}_{U|X}^{(0)})\ \forall n$ and the complete trajectory of the algorithm for a particular initial condition is  contained in $H_{\delta, \lambda}({P}_{U|X}^{(0)})$ which is clearly a path-connected set.  
\begin{lemma}
\label{identidades}
Consider $\lambda\in(0.5,1]$ and ${P}_{U|X}^{(0)}\in G_{\delta,\lambda}$ for a given value\footnote{It is easy to show that we can always find a value of $\delta$ such that this condition is satisfied.} of $\delta$. If the optimal solution ${P}_{U|X}^{*,\lambda}$ lies in $H_{\delta,\lambda}({P}_{U|X}^{(0)})$, the function $f(\lambda,{P}_{U|X})$ is concave in $H_{\delta,\lambda}({P}_{U|X}^{(0)})$, then the following inequalities hold for every $n$:
\begin{align}
V_\lambda&\leq\lambda\sum_{x,y,z,u}P_{U|X}^{*,\lambda}P_{XYZ}\log\left(\frac{Q_{U|YZ}^{(n+1)}}{P_{U|X}^{(n)}}\right)\nonumber\\
&+(2\lambda-1)\sum_{x,z,u}P_{U|X}^{*,\lambda}P_{XZ}\log\left(\frac{Q_{X|ZU}^{(n+1)}}{P_{X|Z}}\right),\label{eq:v1}\\
V_\lambda&-F_\lambda^{(n+1)}\leq\lambda\sum_{x,u}P_{U|X}^{*,\lambda}{P}_{X}\log\left(\frac{{P}_{U|X}^{(n+1)}}{{P}_{U|X}^{(n)}}\right).
\end{align}
\end{lemma}

\begin{IEEEproof}
	See Appendix \hyperref[app_dpos]{B}.
\end{IEEEproof}

\begin{thm}
	\label{convergencia}
	Consider $\lambda\in(0.5,1]$ and ${P}_{U|X}^{(0)}\in G_{\delta,\lambda}$ for a given value of $\delta$. If the optimal solution $P_{U|X}^{*,\lambda}$ lies in $H_{\delta,\lambda}({P}_{U|X}^{(0)})$ and the function $f(\lambda,{P}_{U|X})$ is concave in $H_{\delta,\lambda}({P}_{U|X}^{(0)})$ and ${P}_{U|X}^{(0)}$ is such that $|\text{supp}({P}_{U|X}^{(0)})|=|\mathcal{U}|$, then:
	\begin{enumerate}
		\item Convergence of $F_\lambda$: 
		$
		\lim_{n\rightarrow\infty}F_\lambda^{(n)}=V_\lambda;
		$
		
		\item Convergence of ${P}^{(n)}_{U|X}$:
		$
		\lim_{n\rightarrow\infty}{P}_{U|X}^{(n)}=P_{U|X}^{*,\lambda}.
		$
	\end{enumerate}
\end{thm}

\begin{IEEEproof}
	\begin{enumerate}
		\item For any integer $N\geq 1$, from Lemma \ref{identidades} we can bound: 
		\begin{align}
		&\sum_{n=0}^{N-1}V_\lambda-F_\lambda^{(n+1)}\leq\lambda\sum_{x,u}P_{U|X}^{*,\lambda}{P}_{X}\log\left(\frac{{P}_{U|X}^{(N)}}{{P}_{U|X}^{(0)}}\right)\\
		&=\lambda\left(\mathbb{E}_X\left[\mathcal{D}(P_{U|X}^{*,\lambda}\|{P}_{U|X}^{(0)})\right]-\mathbb{E}_X\left[\mathcal{D}(P_{U|X}^{*,\lambda}\|{P}_{U|X}^{(N)})\right]\right)\\
		&\leq\lambda\mathbb{E}_X\left[\mathcal{D}(P_{U|X}^{*,\lambda}\|{P}_{U|X}^{(0)})\right],\label{ec:sumbound}
		\end{align}
		where the last term is finite because $|\text{supp}({P}_{U|X}^{(0)})|=|\mathcal{U}|$. From claim 2) in Lemma~\ref{fprop} we have: $V_\lambda\geq F_\lambda^{(n+1)}$, and $F_\lambda^{(n)}$ is non-decreasing in $n$. Thus, for $N\rightarrow\infty$, the series converges and $F_\lambda^{(n+1)}\overset{n\rightarrow\infty}{\longrightarrow}V_\lambda$.
		\item From Lemma \ref{fprop}, $F_\lambda^{(n+1)}\geq f(\lambda,{P}_{U|X}^{(n)})\geq F_\lambda^{(n)}$, so using the previous claim $f(\lambda,{P}_{U|X}^{(n)})\overset{n\rightarrow\infty}{\longrightarrow} V_\lambda=f(\lambda,P_{U|X}^{*,\lambda})$. From Lemma \ref{adentro!} we have that ${P}_{U|X}^{(n)}\in H_{\delta,\lambda}({P}_{U|X}^{(0)})$ for all $n$. As $f(\lambda,{P}_{U|X})$ is concave in $H_{\delta,\lambda}({P}_{U|X}^{(0)})$ and its optimal point $P^{*,\lambda}_{U|X}$ is unique, it is easy to check that ${P}_{U|X}^{(n)}\overset{n\rightarrow\infty}{\longrightarrow}P_{U|X}^{*,\lambda}$.
	\end{enumerate}
\end{IEEEproof}
As $f(\lambda,P_{U|X})$ is not globally concave, the $\delta$-superlevel set $G_{\delta,\lambda}$ neither convex but may not also be connected. By Lemma \ref{adentro!}, the algorithm proposed stays in the path-connected component $H_{\delta,\lambda}(P^{(0)}_{U|X})$ which is determined by the initial condition. If the the optimal point $P_{U|X}^{*,\lambda}$ is contained in the right path-connected component $H_{\delta,\lambda}({P}_{U|X}^{(0)})$ and $f(\lambda,P_{U|X})$ is locally concave around the optimal point $P_{U|X}^{*,\lambda}$,  which is something reasonable to expect because of the smoothness of $f(\lambda,P_{U|X})$. The above results give positive answers regarding the convergence of the algorithm to the optimal point $P_{U|X}^{*,\lambda}$. However, to avoid convergence to a local maximum located in a wrong path-connected component $G_{\delta,\lambda}$, a simple solution in practice is to run the algorithm from a few different initial conditions and keep the one that provides the largest value of $F$ after stopping condition is met.

\subsection{Optimal solution and stopping condition}
\label{subsec:stopping}
We now consider some properties of the optimal solution $P^{*}_{U|X}$ and the stopping condition for the proposed algorithm that can be obtained from them. Starting by the next lemma:
\begin{lemma}
	\label{kkt}
	Consider $\lambda\in(0.5,1]$. If $f(\lambda,{P}_{U|X})$ is concave in a vicinity of the optimal solution  $P_{U|X}^{*,\lambda}$, we have
	\begin{equation}
	\begin{array}{cl}\alpha^{*}(\lambda,u)=V_\lambda, &\mbox{for}\ u\in\mathcal{U}\ \mbox{such that}\  P_{U|X}^{*,\lambda}>0\ \mbox{and}\ \forall x\in\mathcal{X}\\
	\alpha^{*}(\lambda,u)\leq V_\lambda, &\mbox{otherwise},\end{array}
	\end{equation}
	where
	\begin{align}
	\alpha^{*}(\lambda,u)&\coloneqq(2\lambda-1)\sum_{(x,z)\in \mathcal{X}\times \mathcal{Z}}{P}_{XZ}\log\left(\frac{{Q}_{X|ZU}^{*,\lambda}}{{P}_{X|Z}}\right)\nonumber\\
	&+\lambda\sum_{(x,y,z)\in \mathcal{X}\times\mathcal{Y}\times \mathcal{Z} }{P}_{XYZ}\log\left(\frac{{Q}_{U|YZ}^{*,\lambda}}{P_{U|X}^{*,\lambda}}\right).
	\end{align}
\end{lemma}
\begin{IEEEproof}
	See Appendix \hyperref[app_kkt]{C}.
\end{IEEEproof}
It is interesting to observe that the optimal solution $P_{U|X}^{*,\lambda}$ is such that for each value of $u\in\mathcal{U}$ where $P_{U|X}^{*,\lambda}>0$ the value of $\alpha^{*}(\lambda,u)$ is constant and equal to the maximum value $V_\lambda$. Similar results are obtained for the optimum solutions for the capacity of a discrete memoryless channel and the rate-distortion function for a discrete memoryless source \cite{gallager}. In particular, we have:
\begin{equation}
V_\lambda=\max_{u\in\mathcal{U}}\alpha^{*}(\lambda,u).
\end{equation}
From these results, we can consider the quantity:
\begin{align}
I_\lambda^{(n+1)} & \coloneqq \max_{u\in\mathcal{U}}\;(2\lambda-1)\sum_{(x,z)\in \mathcal{X}\times \mathcal{Z}}{P}_{XZ}\log\left(\frac{{Q}_{X|ZU}^{(n+1)}}{{P}_{X|Z}}\right)\nonumber\\
&+\lambda\sum_{(x,y,z)\in \mathcal{X}\times\mathcal{Y}\times \mathcal{Z} }{P}_{XYZ}\log\left(\frac{{Q}_{U|YZ}^{(n+1)}}{{P}_{U|X}^{(n)}}\right).
\end{align}
It is clear from (\ref{eq:v1}) that for $\lambda\in(0.5,1]$, $I_\lambda^{(n+1)}\geq V_\lambda$. This suggests that a stopping condition specially matched to the optimal value $P_{U|X}^{*,\lambda}$ could be implemented by checking the condition: $I_\lambda^{(n)}-{F}_{\lambda}^{(n)}\leq\epsilon$ for a sufficiently small $\epsilon>0$.

\section{Numerical Evaluation}
\label{sec:experimentos}
In this section, we apply the proposed algorithm to different application problems.

\subsection{Example of computation of a  relevance-rate region}

\begin{figure}
	\includegraphics[width=0.48\textwidth]{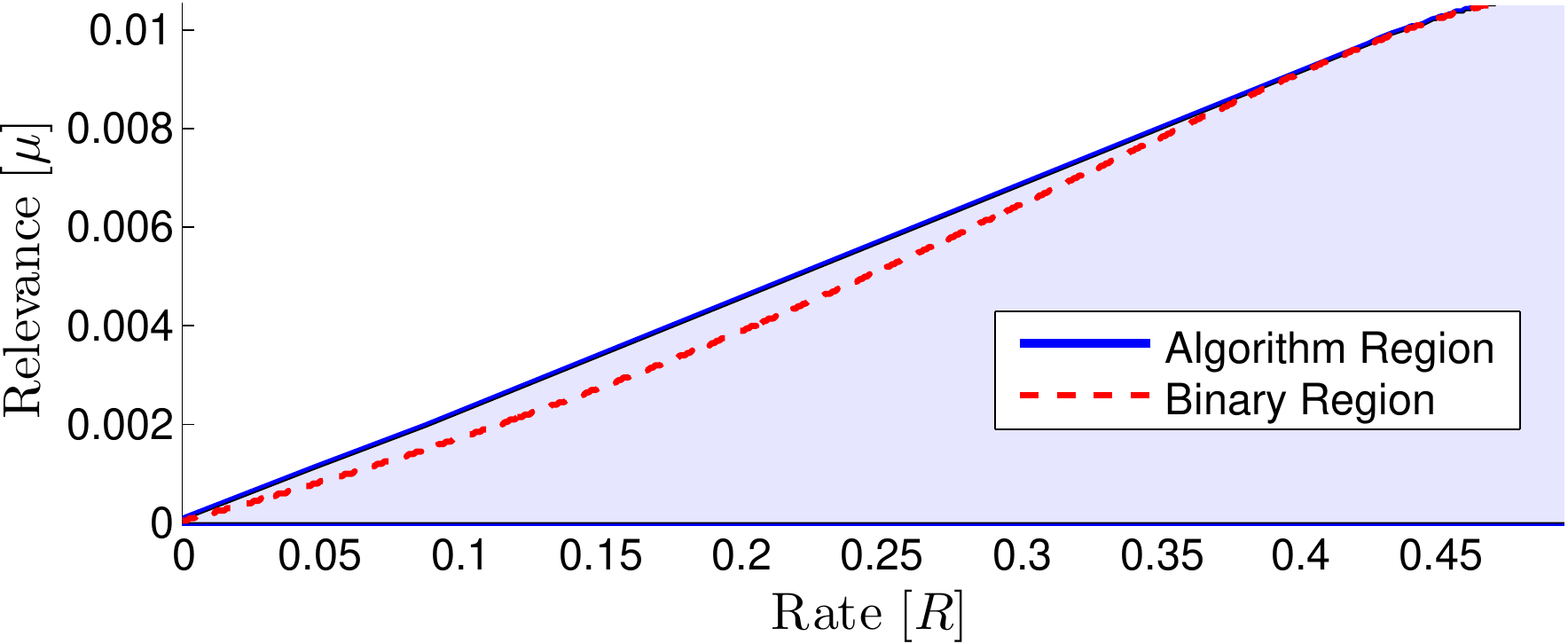}
	\caption{Rate-relevance region corresponding to a double symmetric binary source.}
	\label{fig:ej1}
\end{figure}

According to the discussion presented in Section \ref{sec:problem}, although region $\mathcal{R}$ is convex, the problem of obtaining its upper-boundary (or the relevance-rate function defined in expression \eqref{eq:rel_rate_funct}) is not a convex one. For this reason, only a small number of cases can be solved in closed form. One of them is the double binary source problem with binary side information at the decoder which was completed solved in \cite{2016arXiv160401433V}. In this problem, the pmf $P_{XYZ}$ is a probability measure corresponding to a source $(X,Y,Z)$ such that $Y\mkv X\mkv Z$ and $(X,Z)$ form a doubly symmetric binary source with crossover probability $p$, and $(X,Y)$ forms a doubly symmetric binary source with crossover probability $p$. It was shown in~\cite{2016arXiv160401433V} that
the relevance-rate region is given by the convex hull~\cite{rockafellar_convex_1970} of the following region:
\begin{align}
\mathcal{R}_b\coloneqq   \big\{(R,\mu):R\geq I(X;U|Z),\ \mu\leq I(Y;U|Z),\nonumber\\
\ \ \ \mbox{with}\ P_{U|X}=\text{BSC}(r)\ \ \forall r\in[0,0.5]\big\}, 
\end{align}
where BSC$(r)$ denotes a \emph{binary symmetric channel} with crossover probability $r$. It is clear that the algorithm presented in Section \ref{sec:algoritmo} allows the computation of all relevance-rate pairs in $\mathcal{R}$ for an arbitrary pmf $P_{XYZ}$. This can be easily done by running the algorithm for a sufficient dense grid of points $\lambda\in(0.5,1]$ for the desired pmf $P_{XYZ}$. In order to test the suitability of the algorithm for this task we used it with the source $(X,Y,Z)$ described above setting parameters $p=0.1$ and $q=0.4$. In Fig.~\ref{fig:ej1}, we show the region obtained by our algorithm and the upper-boundary of region $\mathcal{R}_b$. We can observe that the region obtained by the algorithm coincides with the convex hull of $\mathcal{R}_b$.

\subsection{Compression-based regularization learning}

In the previous sections, we have shown that the problem of maximizing the relevance $I(U;Y|Z)$ subject to a mutual-information constraint $I(U;X|Z)\leq R$ is equivalent to that of maximizing $f(\lambda,P_{U|X})$ which introduces the penalization term: $(1-\lambda)I(U;X|Z)$.  We now show that this constraint can act as a regularization when applied to situations where the joint statistics controlling the observations $P_{XYZ}$ is not known but it is estimated from training samples.  Indeed, Shamir~\emph{et al.}~\cite{shamir10} have already showed evidence that this term can help to prevent ``overfitting" and this idea was also exploited in~\cite{tishby15,tishby17} to justify some features of deep learning algorithms. It should mentioned that these analysis were performed for the classical IB method without the presence of side information. In this section, we provide numerical evidence that the desired regularization effects hold  in our multi-task learning setup. 

Consider a multi-task supervised classification problem and define the average cross-entropy risk:
\begin{equation}
\textrm{Risk}\big(P_{U|X},P_{\hat{Y}|UZ}\big)\coloneqq   \mathbb{E}_{{P}_{XYZ}P_{U|X}}[- \log  P_{\hat{Y}|UZ}(Y|UZ) ]\label{eq-risk}
\end{equation}
with respect to $P_{U|X}$ and $P_{\hat{Y}|UZ}$. We notice that this risk  is not necessarily equivalent to the classification error. However, it is easy to check that it is an appropriate surrogate: 
\begin{equation}
\mathbb{P}\big(Y\neq\hat{Y}\big)\leq1-2^{-\textrm{Risk}\big(P_{U|X},P_{\hat{Y}|UZ}\big)}.
\end{equation}
\begin{figure}[t]
	\centering
	\includegraphics[width=0.48\textwidth]{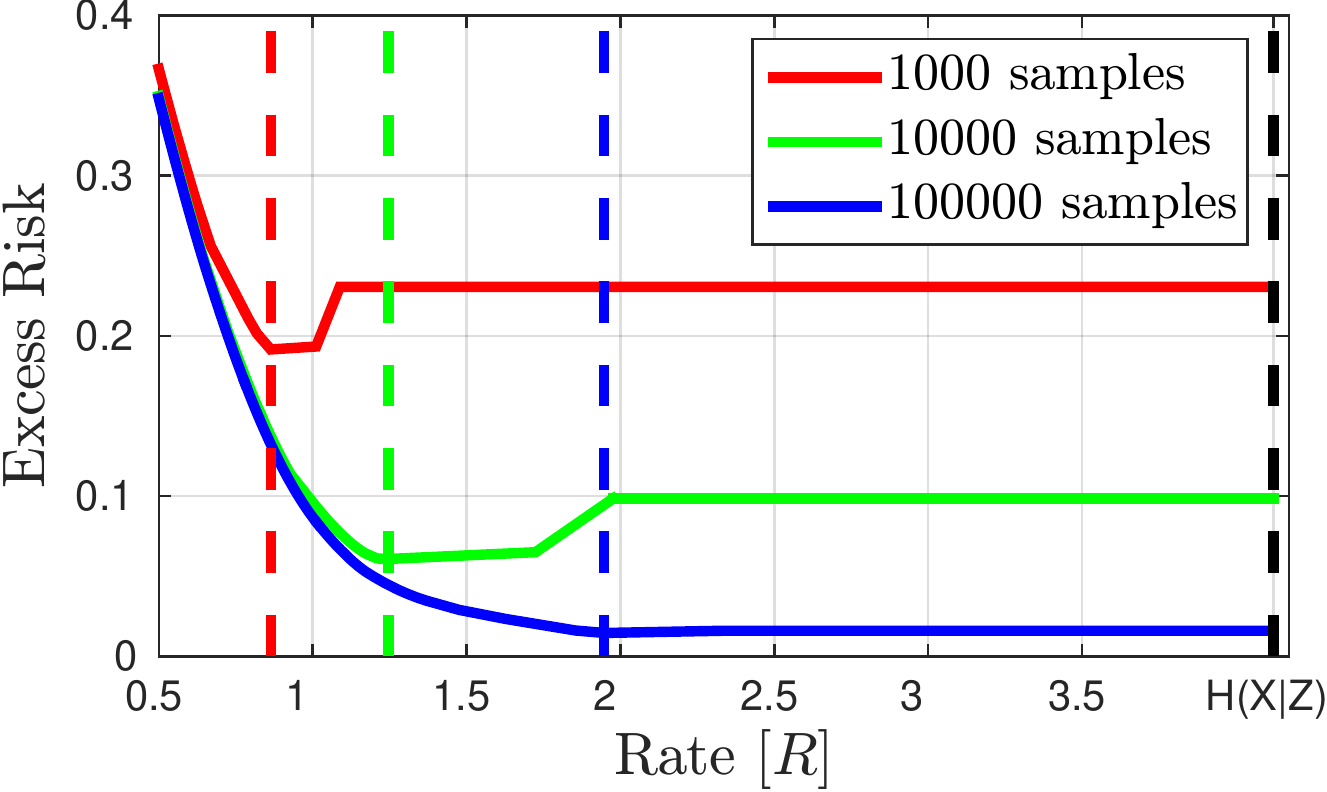}
	\caption{Excess risk~\eqref{eq-excess-risk} as a function of the information rate.}
	\label{fig:regular}
\end{figure}
Finding the optimal encoder in \eqref{eq-risk}  requires knowledge of the underlying distribution $P_{XYZ}$. From a practical perspective, as the input to  the proposed algorithm, we will use the data sampling distribution $\hat{P}_{XYZ}$ based on $n$ training labeled examples. By introducing the rate constraint (or penalization), the optimization problem is reduced to optimizing $L(R,\hat{P}_{XYZ})$ in~\eqref{eq:rel_rate_funct} from which the resulting  encoder $\hat{P}_{U|X}^{*,\lambda}$ is derived while the decoder $\hat{P}_{\hat{Y}|UZ}^{*,\lambda}$ follows from expression\footnote{Notice that when the decoder is chosen as in (\ref{ec:decoder}), then $\textrm{Risk}\big(P_{U|X},P_{\hat{Y}|UZ}\big)=H(Y|UZ)\geq H(Y|XZ)$, where the inequality is a consequence of the Markov chain $U\mkv X\mkv (Y,Z)$.} (\ref{ec:decoder}). The measure of merit will be the \emph{Excess-risk}: 
\begin{equation}
\textrm{Excess-risk} \coloneqq  \textrm{Risk}\big(\hat{P}_{U|X}^{*,\lambda},\hat{P}_{\hat{Y}|UZ}^{*,\lambda}\big) - H(Y|XZ) \label{eq-excess-risk}
\end{equation}
that is the difference between the minimum Bayesian risk $H(Y|UZ)$ and the risk induced from the suboptimal encoder $\hat{P}_{U|X}^{*,\lambda}$ obtained by optimizing w.r.t the sample distribution $\hat{P}_{XYZ}$ subject to the rate constraint.\par
Experiments will be performed by using synthetic data with alphabets $|\mathcal{X}|=64$, $|\mathcal{Y}|=4$, $|\mathcal{Z}|=2$. The random variable $Z$ is assumed to follow a \emph{Bernoulli} distribution with random parameter $p\in[0,1]$ while the joint distribution $P_{XY|Z=z}$ is defined as a  \emph{Restrict Boltzmann Machine} (see~\cite{38136} for further details) with parameters randomly drawn for each $z\in\mathcal{Z}$. 

In Fig. \ref{fig:regular}, we plot the excess risk curve as a function of the rate constraint for different size of training samples. With dash lines we denoted the rate for which the excess risk achieves its minimum. When the number of training samples increases the optimal rate $R$ approaches its maximum possible value: $H(X|Z)$ (dashed in black). We emphasize that for every curve there exists a different limiting rate $R_{\rm lim}$, such that for each $R\geq R_{\rm lim}$, the excess-risk remains constant with value $\hat{I}(X;Y|Z)$. It is not difficult to check that $R_{\rm lim}=\hat{H}(X|Z)$. Furthermore, for every size of training samples, there is an optimal value of $R_{\rm opt}$ which provides the lowest excess-risk in~\eqref{eq-excess-risk}. In a sense, this is indicating that the rate $R$ can be interpreted as an effective regularization term and thus, it can provide robustness for learning in practical scenarios in which the true input distribution is not known and the empirical data distribution is used. It is worth to mention that when more data is available then the optimal value of the regularizing rate $R$  becomes less critical. Of course, this fact was expected since as the amount of training data increases the empirical distribution approaches the true data-generating distribution. 

\subsection{Hierarchical text categorization}
The high dimensionality of texts can become a severe deterrent in applying complex learners like support vector machines (SVM)~\cite{38136} to the task of text classification. Word clustering is a powerful alternative to feature selection for reducing the dimensionality of text \cite{slonim2001,dhillon03}. This issue can be alleviated by intelligently grouping different classes in disjoint sub-categories. In this way, a first  classification problem can be set over the generated sub-categories and the information extracted can be used in a second classification problem to discriminate better between classes. This is the case in hierarchical text classification \cite{chechik02,vinokouro02}. 
\begin{figure}[!t]
	\begin{center}
		\pgfdeclarelayer{background}
		\pgfdeclarelayer{foreground}
		\pgfsetlayers{background,main,foreground}
		\tikzstyle{daemon}=[draw, fill=blue!20, text width=2.5em, text centered, drop shadow]
		\tikzstyle{dots} = [above, text centered]
		\tikzstyle{wa} = [daemon, fill=blue!10, minimum height=2.5em, rounded corners, drop shadow, inner sep=0.1em]
		\tikzstyle{arrow}=[draw, -latex] 
		\def\blockdist{1.3}
		\def\edgedist{1.5}
		\begin{tikzpicture}
		\node (e1) [wa,text width=2.0cm] {Encoder $1$};
		\path (e1.south)+(0.0,-1.5) node (e2)[wa, text width=2.0cm] {Encoder $2$};
		\path (e1.east)+(2.0,0.0) node (d1)[wa, text width=2.0cm] {Decoder $1$};
		\path (e2.east)+(2.0,0.0) node (d2)[wa, text width=2.0cm] {Decoder $2$};
		
		\path (e1.west)+(-0.5,-0.12) node (fan1)[dots] {};
		\path (e1.west)+(-1.2,-0.24) node (x)[dots] {$X$};
		\path (d1.east)+(1.2,-0.335) node (y1)[dots] {$P_{\hat{Y}_1|U_1}$};
		\path (d2.east)+(1.35,-0.334) node (y2)[dots] {$P_{\hat{Y}_2|U_1U_2}$};
		\path (e1.east)+(0.46,-1.0) node (fan2)[dots] {};
		
		\path [draw, ->] (x.east) -- node [above] {} (e1.west);
		\path [draw, ->] (fan1.center) |- node [above] {} (e2.west);
		\path [draw, ->] (d1.east) -- node [above] {} (y1.west);
		\path [draw, ->] (d2.east) -- node [above] {} (y2.west);
		\path [draw, ->] (e1.east) -- node [above] {$U_1$} (d1.west);
		\path [draw, ->] (e2.east) -- node [above] {$U_2$} (d2.west);
		\path [draw, -] (e1.east) -| node [above] {} (fan2.center);
		\path [draw, ->] (fan2.center) -| node [above] {} (d2.north);
		
		\draw [green,line width=0.3mm,dashed] (-1.4,-0.7) rectangle (4.4,0.7);
		\draw [red,line width=0.3mm,dashed] (-1.4,-2.7) rectangle (4.4,-1.1);
		\end{tikzpicture}
	\end{center}
	\caption{Hierarchical text categorization and multi-task learning.}
	\label{fig:db_tc}
\end{figure}
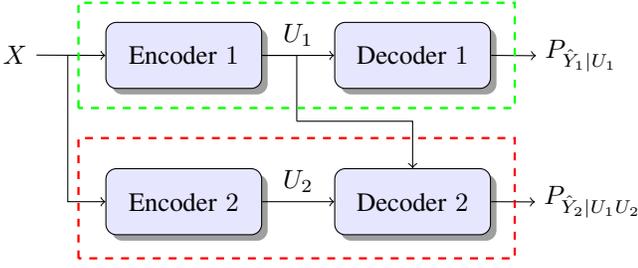 
We approach this problem based on the scheme of Fig. \ref{fig:db_tc}. Consider a document $d$ consisting of different words $X$. We want to estimate the class $Y_2$ to which the document belongs by using information related to a sub-category $Y_1$ (typically related to the text topic) to which the same document also belongs. To this end, assume a pair of encoder 1-decoder 1 infers  the document sub-category $\hat{Y}_1$ by using our algorithm without side information (i.e. $Z$ is a degenerate RV) and with input $P_{XY_1}$. This is clearly a standard classification problem where $U_1$ is the feature that encoder 1 extracts from $X$. Encoder 2-decoder 2 pair generates the final classification in $\hat{Y}_2$ by using the algorithm with input $P_{XY_1U_1}$. $U_1$ can be considered as side information available at decoder 2. This problem can be interpreted as a MTL problem where the different classification tasks to be inferred  by decoder 2 are induced by the features extracted from encoder 1. 

Assume a training set consisting of documents belonging to $|\mathcal{Y}_2|$ classes, which has $|\mathcal{X}|$ different words. The distribution $P_{Y_1|Y_2}$ is known because the sub-category $Y_1$ is a deterministic function of the more refined class $Y_2$ (i.e. $Y_1=h(Y_2)$). The class priors $P_{Y_2}$ are replaced by the empirical distribution and the words distribution conditional to the class, $P_{X|Y_2}$ is estimated using Laplace rule of succession \cite{bishop_praml}. Imposing the Markov chain $U_1\mkv X\mkv Y_2$, the resulting joint pmfs are given by $P_{XY_2U_1}=P_{U_1|X}P_{X|Y_2}P_{Y_2}$ and 
\begin{equation}
P_{XY_1}=\sum_{y_2\in\mathcal{Y}_2}P_{Y_1|Y_2}P_{X|Y_2}P_{Y_2}.
\end{equation}
\begin{figure}[!t]
	\centering 
	\includegraphics[width=0.48\textwidth]{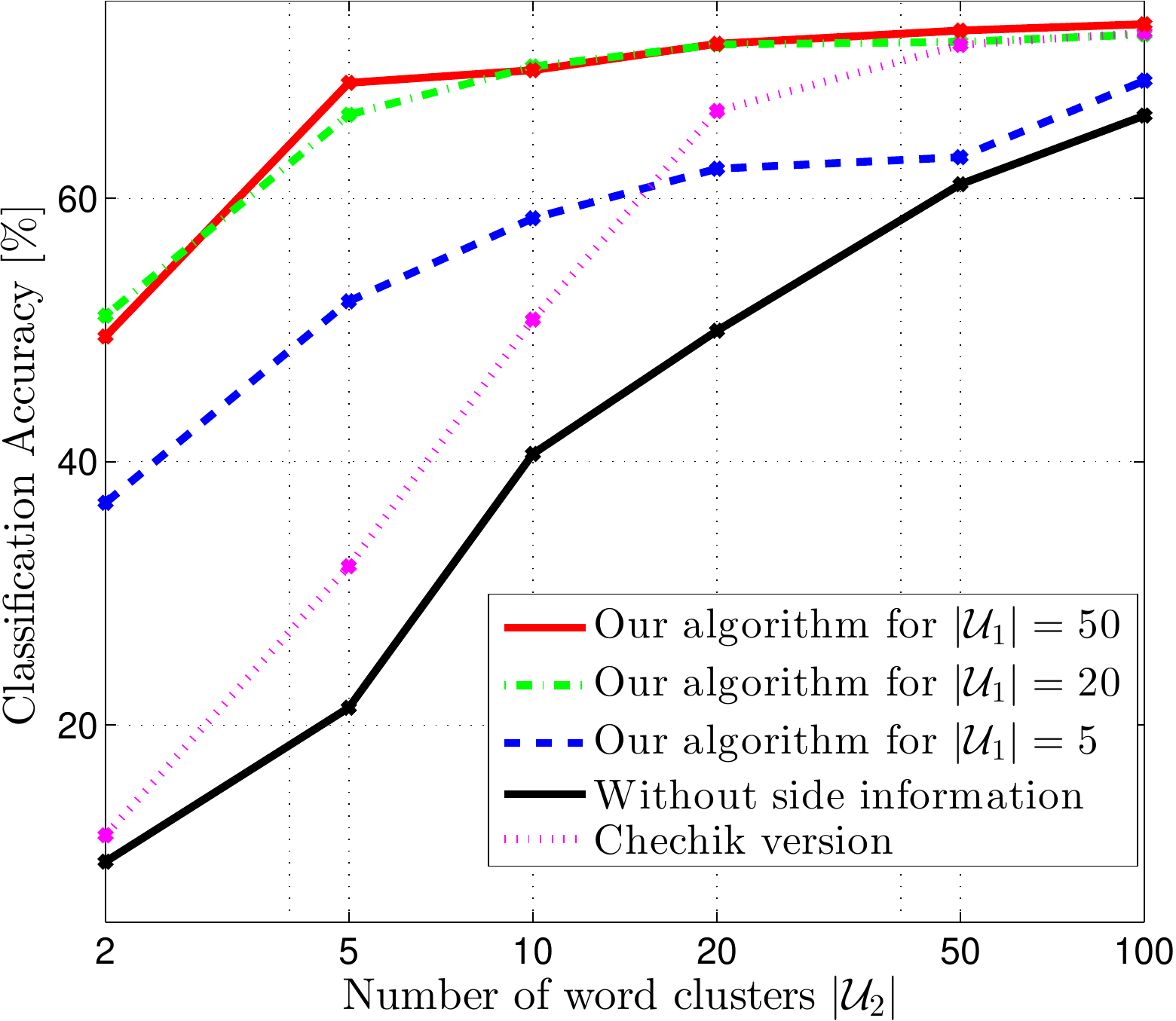}
	\caption{Classification Accuracy in the hierarchical text categorization problem.}
	\label{fig:y2}
\end{figure}
Once pmfs $P_{U_1|X}$ and $P_{U_2|X}$ are calculated using the proposed algorithm, we estimate the class of the document $\hat{y}_2(d)$. Assuming a generative multinomial model, and conditional independence between clusters, the maximum a posteriori probability, which computes the most probable class for a document $d$, is given by (see  \cite{dhillon03} for details):
\begin{align}
&\hat{y}_2(d)= 
\arg\max_{y\in\mathcal{Y}_2}P_{Y_2}\prod_{u_1,u_2}\left(P_{U_1U_2|Y_2}\right)^{n(u_1,u_2,d)}\\
&\;\equiv \arg\max_{y\in\mathcal{Y}_2}\log \left(P_{Y_2}\right)+\sum_{u_1,u_2}{n(u_1,u_2,d)}\log P_{U_1U_2|Y_2},
\end{align}
where 
\begin{equation}
P_{U_1U_2|Y_2}=\frac{\sum_xP_{XY_2U_1}P_{U_2|X}}{P_{Y_2}},
\end{equation}
and $n(u_1,u_2,d)$ is the number of jointly occurrences of clusters $(u_1,u_2)$ in the document $d$ computed with:
\begin{equation}
u_1(x)\coloneqq  \arg\max_u P_{U_1|X}(u|x),\; u_2(x)\coloneqq  \arg\max_u P_{U_2|X}(u|x).
\end{equation}

We test the above proposed classification procedure on the 20 Newsgroups (20Ng) dataset \cite{lang95}. This contains 11269 documents for training and 7505 for testing evenly divided among 20 UseNet Discussion groups or classes. Each newsgroup represents one class in the classification task. The train dataset had 53975 different words. The $20$ Newsgroups correspond to $6$ topics. The sub-category $Y_1$ represents  the topic among $6$ possibilities  and the refined classification $Y_2$ is the class among $20$ possibilities.\par
In Fig. \ref{fig:y2}, our algorithm performance ($\lambda=0.99$) versus $|\mathcal{U}_2|$ is compared with the algorithm without side information (which is a single-task setup) and the one proposed in \cite{chechik02}. It is interesting to mention that the single-task setting and the one in \cite{chechik02} can be covered using the proposed algorithm. In particular, for the single-task setup, we estimate the final class with $P_{XY_2}$ as the input of the proposed algorithm ($Z$ is a degenerate RV). Our setting and the one in \cite{chechik02} show an improvement with respect to the single task setup  without side information. This suggests that exploiting the common features in MTL may be advantageous. With $|\mathcal{U}_1|=20$ and $|\mathcal{U}_2|=5$, our method achieves $66.36\%$ of accuracy. For which we exploit the additional information in a structured manner to show an improvement with respect to the other proposals.

\begin{remark}
	There exists a strong relationship between our objective and the one in \cite{chechik02} when we redefine the tasks as the classification within the different sub-categories. In this case, referring to Fig. \ref{fig:db2} we consider $Y=Y_2$ and the side information as a deterministic function of the task $Z=g(Y)$ (every class in the same sub-category belongs to the same $Z$). Our objective $f(\lambda,P_{U|X})$ can be written as:
	\begin{align}
	&f(\lambda,P_{U|X})=\lambda I(Y;U|g(Y))-\bar{\lambda}I(X;U|g(Y))\\
	&=\lambda\left[I(Y;U)-I(g(Y);U)\right]-\bar{\lambda}\left[I(X;U)-I(g(Y);U)\right]\\
	&=\lambda I(Y;U)-(2\lambda-1)I(g(Y);U)-\bar{\lambda}I(X;U),\label{eq:chechik}
	\end{align} 
	where $\bar{\lambda}=1-\lambda$ and $f(\lambda,P_{U|X})$ depends on the source via the marginal distributions: $P_{XY}$ and $P_{XZ}$. This expression is the cost function proposed in \cite{chechik02} with $\beta=\frac{\lambda}{1-\lambda}$ and $\gamma=\frac{2\lambda-1}{\lambda}$.
\end{remark}

\section{Conclusions}
\label{sec:conclusiones}

From information-theoretic methods, we have investigated the supervised learning framework of Multi-task learning in which an encoder builds a common representation intended to several related tasks. We derived an iterative learning algorithm from the principle of compression-based regularization that uses compression  as a natural safeguard against overfitting. Numerical evidence showed that there exists an optimal compression rate minimizing the \emph{excess risk} according to the amount of available training data. Indeed, this rate increases  with the size of the training set. An application to hierarchical text categorization was also considered. 

At present, several open questions remain regarding the statistical regularization properties of building compact representations of data. It is clear that both further theoretical and practical studies are required. Applications of our  algorithm to other multi-task learning setups, besides the hierarchical text categorization one, should also deserve additional efforts.

\appendix

\section*{Appendix A: Proof of Lemma \ref{lem:cvx}}
\label{app_cvx}

We show that the region $\mathcal{R}$ is closed, convex and we bound the cardinality of the RV $U$. A region $\mathcal{R}$ is closed iff it contains the limit of every converging sequence whose terms lie in $\mathcal{R}$. Let $(\mu^{(k)},R^{(k)})\in\mathcal{R}$ such that $(\mu^{(k)},R^{(k)})\overset{k\rightarrow\infty}{\longrightarrow}(\mu,R)$. As $(\mu^{(k)},R^{(k)})\in\mathcal{R}$ we have that exists $P_{U|X}^{(k)}$ such that:
\begin{equation}
R^{(k)}\geq I^{(k)}(X;U|Z),\;\;\mu^{(k)}\leq I^{(k)}(Y;U|Z).
\end{equation}
We have then a sequence of conditional probability distributions $\left\{P_{U|X}^{(k)}\right\}_{k=1}^{\infty}$. As this sequence is in a compact set (i.e. the set of all conditional PDs with alphabets $\mathcal{X}$ and $\mathcal{U}$) it should exists a converging subsequence $P_{U|X}^{(k_n)}$ with $n\in\mathbb{N}$ and limiting  point $P_{U|X}^{\rm lim}$. Consider the subsequence $(\mu^{(k_n)},R^{(k_n)})$, it is straightforward to check  that for any $\epsilon>0$ and $n$ large enough, $|R-R^{(k_n)}|<\epsilon$ and $|\mu-\mu^{(k_n)}|<\epsilon$. Then, we have:
\begin{equation}
R\geq I^{\rm lim}(X;U|Z)-\epsilon,\;\;\mu\leq I^{\rm lim}(Y;U|Z)+\epsilon.
\end{equation}
As $\epsilon$ is arbitrary we conclude that $(\mu,R)\in\mathcal{R}$.\par 
For the convexity analysis consider $(R_1,\mu_1),(R_2,\mu_2)\in\mathcal{R}$. We will prove $\theta(R_1,\mu_1)+(1-\theta)(R_2,\mu_2)\in\mathcal{R}$ $\forall$ $\theta\in[0,1]$.\par 
If $(R_1,\mu_1)\in\mathcal{R}$, $R_1\geq I(X;V_1|Z)$ and $\mu_1\leq I(Y;V_1|Z)$ for a conditional distribution ${P}_{V_1|X}$. Similarly, if $(R_2,\mu_2)\in\mathcal{R}$, $R_2\geq I(X;V_2|Z)$ and $\mu_2\leq I(Y;V_2|Z)$ with distribution ${P}_{V_2|X}$. We define $T\sim\text{Ber}(\theta)$ (Bernoulli RV with parameter $\theta$) independent of $(X,Y,Z,V_1,V_2)$, and the RV:
\begin{equation}
V=\left\{\begin{array}{cc}V_1,&\text{If  }\;T=1\\V_2,&\text{If  }\;T=0. \end{array}\right.
\end{equation}
Then, 
\begin{align}
\theta& R_1+(1-\theta)R_2\nonumber\\
&\geq\theta I(X;V_1|Z)+(1-\theta)I(X;V_2|Z)\\
&=\theta I(X;V|Z,T=1)+(1-\theta)I(X;V|Z,T=0)\\
&=I(X;V,T|Z)\\
&=I(X;U|Z),
\end{align} 
where $U=VT$. The same happens with the relevance:
\begin{align}
\theta&\mu_1+(1-\theta)\mu_2\nonumber\\
&\leq\theta I(Y;V_1|Z)+(1-\theta)I(Y;V_2|Z)\\
&=\theta I(Y;V|Z,T=1)+(1-\theta)I(Y;V|Z,T=0)\\
&=I(Y;V,T|Z)\\
&=I(Y;U|Z).
\end{align}   
It is necessary to show that $U$ have the Markov property $U\mkv X\mkv (Y,Z)$:
\begin{align}
I&(Y,Z;V,T|X)=I(Y,Z;V|X,T)\\
&\;=\theta I(Z;V|X,T=1)+(1-\theta)I(Z;V|X,T=0)\\
&\;=\theta I(Z;V_1|X)+(1-\theta)I(Z;V_2|X)=0.
\end{align}
It is clear then that $\mathcal{R}$ is convex. Now, we will show that the cardinality of RV $U$ can be bounded $|\mathcal{U}|\leq|\mathcal{X}|+1$ without loss of generality. This follows easily from the Support Lemma \cite[app.~C]{gamal}, witch follows from Carath\'eodry Theorem.
\begin{lemma}[Support Lemma]
\label{caratheodry}
Let $d$ functions $g_1,g_2,\dots,g_d$ of conditional probabilities $P_{X|U}$. Then, for all $U$ exists $U^\prime$ with cardinality $|\mathcal{U}^\prime|\leq d$ s.t. $\mathbb{E}_U[g(P_{X|U})]=\mathbb{E}_{U^\prime}[g(P_{X|U^\prime})]$.
\end{lemma} 

For our problem the choice of these functions is done in order to preserve \emph{Markov Chains} and the mutual information expressions for the bounds on the rate and the relevance. It is an almost trivial exercise to check that the functions amount to a quantity of $|\mathcal{X}|+1$ atoms. 

\section*{Appendix B: Proof of Lemma \ref{identidades}}
\label{app_dpos}
In order to show Lemma~\ref{identidades}, we will need the following auxiliary result:
\begin{lemma}
	\label{Dpos}
	Let $\lambda\in(0.5,1]$ and $T$ be a convex set of conditional distributions ${P}_{U|X}$ such that the function $f(\lambda,{P}_{U|X})$ is concave in the domain $T$. Then, $\mathcal{L}[P_a,P_b]\geq0$ for any $P_a,P_b\in T$, where
\begin{align}
\mathcal{L}[P_a,P_b]=&\sum_{(x,y,z,u)}P_a{P}_{X,Y,Z}\big[\lambda \mathcal{D}\left(P_a\|P_b\right)\nonumber\\
&-\lambda \mathcal{D}\left({Q}_{U|Y,Z}[P_a]\|{Q}_{U|YZ}[P_b]\right)\nonumber\\
&-(2\lambda-1)\mathcal{D}\left({Q}_{X|UZ}[P_a]\|{Q}_{X|U,Z}[P_b]\right)\big],
\end{align}
and we have defined, for $i=\{a,b\}$:
\begin{equation}
{Q}_{U|YZ}[P_i]=\sum_{x\in\mathcal{X}}P_i{P}_{X|YZ},\;\;{Q}_{X|ZU}[P_i]=\frac{P_i{P}_{X|Z}}{\sum_xP_i{P}_{X|Z}}.
\end{equation}
\end{lemma}
\begin{IEEEproof}
We start calculating $\frac{\partial f(\lambda,{P}_{U|X})}{\partial{P}_{U|X}}$. For $(u,x)$ such that ${P}_{U|X}=0$ the derivative is zero. For $(u,x)$ such that ${P}_{U|X}>0$, we use the identity: $[f(x)\log(f(x))]^{\prime}=f^{\prime}(x)\log\left(ef(x)\right)$ and obtain:
\begin{align}
&\frac{\partial f(\lambda,{P}_{U|X})}{\partial{P}_{U|X}}=(2\lambda-1){P}_{X}\log\left(e{P}_{U|X}\right)\nonumber\\
&\quad-(2\lambda-1)\sum_z{P}_{X,Z}\log\left(e{P}_{U|Z}\right)\nonumber\\
&\quad-\lambda\;{P}_{X}\log\left(e{P}_{U|X}\right)+\lambda\sum_{y,z}{P}_{XYZ}\log\left(e{P}_{U|YZ}\right)\\
&\;=(2\lambda-1)\sum_z{P}_{XZ}\log\left(\frac{{P}_{U|X}}{{P}_{U|Z}}\right)\nonumber\\
&\quad-\lambda\sum_{y,z}{P}_{XYZ}\log\left(\frac{{P}_{U|X}}{{P}_{U|YZ}}\right)\\
&\;=(2\lambda-1)\sum_z{P}_{XZ}\log\left(\frac{{P}_{X|ZU}}{{P}_{X|Z}}\right)\nonumber\\
&\quad-\lambda\sum_{y,z}{P}_{XYZ}\log\left(\frac{{P}_{U|X}}{{P}_{U|YZ}}\right).\label{derivadaf}
\end{align}
Note that
\begin{equation}
\label{igualdadkkt}
f(\lambda,{P}_{U|X})=\sum_{x,u}{P}_{U|X}\frac{\partial f(\lambda,{P}_{U|X})}{\partial{P}_{U|X}}.
\end{equation}
Then,
\begin{equation}
\sum_{x,u}P_b\left.\frac{\partial f(\lambda,{P}_{U|X})}{\partial{P}_{U|X}}\right|_{P_b}=f(\lambda,P_b).
\end{equation}
Now, let us consider:
\begin{align}
&\sum_{x,u}P_a\left.\frac{\partial f(\lambda,{P}_{U|X})}{\partial{P}_{U|X}}\right|_{P_b}\nonumber\\
&=f(\lambda,P_a)-(2\lambda-1)\sum_{x,z,u}P_a{P}_{XZ}\log\left(\frac{{Q}_{X|ZU}[P_a]}{{Q}_{X|ZU}[P_b]}\right)\nonumber\\
&\;+\lambda\sum_{x,u}P_a{P}_{X}\log\left(\frac{P_a}{P_b}\right)\nonumber\\
&\;-\lambda\sum_{x,y,z,u}P_a{P}_{XYZ}\log\left(\frac{{Q}_{U|YZ}[P_a]}{{Q}_{U|YZ}[P_b]}\right)\\
&=f(\lambda,P_a)\nonumber\\
&\;+\sum_{x,y,z,u}P_a{P}_{XYZ}\left[-(2\lambda-1)\mathcal{D} \left({Q}_{X|ZU}[P_a]\|{Q}_{X|ZU}[P_b]\right)\right.\nonumber\\
&\;\left.+\;\lambda \mathcal{D}\left(P_a\|P_b\right)-\lambda \mathcal{D}\left({Q}_{U|YZ}[P_a]\|{Q}_{U|YZ}[P_b]\right)\right]\\
&=f(\lambda,P_a)+\mathcal{L}[P_a,P_b].
\end{align}
Then,
\begin{align}
&\mathcal{L}[P_a,P_b]=\sum_{x,u}P_a\left.\frac{\partial f(\lambda,{P}_{U|X})}{\partial{P}_{U|X}}\right|_{P_b}-f(\lambda,P_a)\\
&=\sum_{x,u}(P_a-P_b)\left.\frac{\partial f(\lambda,{P}_{U|X})}{\partial{P}_{U|X}}\right|_{P_b}-f(\lambda,P_a)+f(\lambda,P_b).
\end{align}
If $f(\lambda,{P}_{U|X})$ is concave in $T$, then:
\begin{equation}
f(\lambda,P_a)\leq f(\lambda,P_b)+\left.\sum_{x,u}\frac{\partial f(\lambda,{P}_{U|X})}{\partial{P}_{U|X}}\right|_{P_b}\left(P_a-P_b\right),
\end{equation}
and thus: $\mathcal{L}[P_a,P_b]\geq f(\lambda,P_a)-f(\lambda,P_a)=0$.
\end{IEEEproof}

Now we can proceed to the proof of  Lemma \ref{identidades}. In order to show \eqref{eq:v1}, we define the quantity $\mathcal{B}$:
\begin{align}
\mathcal{B}&\coloneqq  (2\lambda-1)\sum_{x,z,u}P_{U|X}^{*,\lambda}{P}_{XZ}\log\left(\frac{{Q}_{X|ZU}^{(n+1)}}{P_{X|Z}}\right)\nonumber\\
&\qquad+\lambda\sum_{x,y,z,u}P_{U|X}^{*,\lambda}{P}_{XYZ}\log\left(\frac{{Q}_{U|YZ}^{(n+1)}}{{P}_{U|X}^{(n)}}\right).
\end{align}
Then, we can write:
\begin{align}
V_\lambda&=(2\lambda-1)\sum_{x,z,u}P_{U|X}^{*,\lambda}{P}_{XZ}\log\left(\frac{{Q}_{X|ZU}^{*}}{P_{X|Z}}\right)\nonumber\\
&\qquad+\lambda\sum_{x,y,z,u}P_{U|X}^{*,\lambda}{P}_{XYZ}\log\left(\frac{{Q}_{U|YZ}^{*}}{P_{U|X}^{*,\lambda}}\right)\\
&=\mathcal{B}+(2\lambda-1)\sum_{x,z,u}P_{U|X}^{*,\lambda}P_{XZ}\mathcal{D}\left({Q}_{X|ZU}^{*}\|{Q}_{X|ZU}^{(n+1)}\right)\nonumber\\
&\qquad+\lambda\sum_{y,z}P_{YZ} \mathcal{D} \left({Q}_{U|YZ}^{*}\|{Q}_{U|YZ}^{(n+1)}\right)\nonumber\\
&\qquad-\lambda\sum_{x}P_X \mathcal{D} \left(P_{U|X}^{*,\lambda}\|{P}_{U|X}^{(n)}\right)\\
&=\mathcal{B}-\mathcal{L}[P_{U|X}^{*,\lambda},{P}_{U|X}^{(n)}].
\end{align}	
Consider an integer $n\geq 1$ and the set $\tilde{G}_{\delta,\lambda}^n$ from the proof of Lemma \ref{adentro!}. It is known that this set is convex and from its definition should contain ${P}_{U|X}^{(n)}$ and the optimal solution $P_{U|X}^{*,\lambda}$. As the function $f(\lambda,P_{U|X})$ is concave in $H_{\delta,\lambda}({P}_{U|X}^{(0)})$ and $\tilde{G}_{\delta,\lambda}^n\subseteq H_{\delta,\lambda}({P}_{U|X}^{(0)})$, we can apply Lemma \ref{Dpos} to $\mathcal{L}[P_{U|X}^{*,\lambda},{P}_{U|X}^{(n)}]$ and conclude that $V_\lambda\leq\mathcal{B}$. We also define:
\begin{align}
\gamma^{(n+1)}_{U|X}=\exp&\left\{\frac{2\lambda-1}{\lambda}\sum_{z}{P}_{Z|X}\log({Q}_{X|ZU}^{(n+1)})\right.\nonumber\\
&\qquad\left.+\sum_{y,z}{P}_{Y|XZ}P_{Z|X}\log({Q}_{U|YZ}^{(n+1)})\right\},
\end{align}
from which it is clear that $P_{U|X}\propto\gamma^{(n+1)}_{U|X}$. It is not hard to see that:
\begin{equation}
\mathcal{B}=\lambda\sum_{x,u}P_{U|X}^{*,\lambda}{P}_{X}\log\left(\frac{\gamma^{(n+1)}_{U|X}}{{P}_{U|X}^{(n)}}\right)+(2\lambda-1)H(X|Z).
\end{equation}
On the other hand, $F_\lambda^{(n+1)}$ can be written as:
\begin{align}
&F_\lambda^{(n+1)}=(2\lambda-1)\sum_{x,z,u}P_{U|X}^{(n+1)}{P}_{XZ}\log\left(\frac{{Q}_{X|ZU}^{(n+1)}}{{P}_{X|Z}}\right)\nonumber\\
&\;-\lambda\sum_{x,y,z,u}P_{U|X}^{(n+1)}{P}_{XYZ}\log\left(\frac{\gamma_{U|X}^{(n+1)}}{{Q}_{U|YZ}^{(n+1)}\sum_{u^\prime}\gamma_{U^\prime|X}^{(n+1)}}\right)\\
&=(2\lambda-1)\sum_{x,z,u}P_{U|X}^{(n+1)}{P}_{XZ}\log\left(\frac{{Q}_{X|ZU}^{(n+1)}}{{P}_{X|Z}}\right)\nonumber\\
&\;+\lambda\sum_{x,y,z,u}P_{U|X}^{(n+1)}{P}_{XYZ}\log\left({Q}_{U|YZ}^{(n+1)}\sum_{u^\prime}\gamma_{U^\prime|X}^{(n+1)}\right)\nonumber\nonumber\\
&\;-(2\lambda-1)\sum_{x,z,u}P_{U|X}^{(n+1)}{P}_{XZ}\log({Q}_{X|ZU}^{(n+1)})\nonumber\\
&\;-\lambda\sum_{x,y,z,u}P_{U|X}^{(n+1)}{P}_{XYZ}\log({Q}_{U|YZ}^{(n+1)})\\
&=\lambda\sum_{x}{P}_{X}\log\left(\sum_{u^\prime}\gamma_{U^\prime|X}^{(n+1)}\right)+(2\lambda-1)H(X|Z).
\end{align}	
Finally,
\begin{align}
V_\lambda-F_\lambda^{(n+1)}&\leq \mathcal{B}-\lambda\sum_{x}{P}_{X}\log\left(\sum_{u^\prime}\gamma_{U^\prime|X}^{(n+1)}\right)\nonumber\\
&\qquad-(2\lambda-1)H(X|Z)\\
&=\lambda\sum_{x,u}{P}_{U|X}^{*,\lambda}{P}_{X}\log\left(\frac{P^{(n+1)}_{U|X}}{{P}_{U|X}^{(n)}}\right).
\end{align}

\section*{Appendix C: Proof of Lemma \ref{kkt}}
\label{app_kkt}
The proofs is along the lines of Karush-Kuhn-Tucker (KKT) conditions \cite{boyd}. In this case we look for the maximum of $f(\lambda,{P}_{U|X})$ subject to $\sum_u{P}_{U|X}=1$ for all $x\in\mathcal{X}$ and ${P}_{U|X}\geq0$ for all $(u,x)\in\mathcal{U}\times\mathcal{X}$. Provided that $f(\lambda,{P}_{U|X})$ is concave in a vicinity of  ${P}_{U|X}^{*,\lambda}$ a necessary condition for the local optimality of ${P}_{U|X}^{*,\lambda}$ is the existence of values $\phi_{x,u},\kappa_x$ such that
\begin{enumerate}
	\item $\frac{\partial f(\lambda,{P}_{U|X}^{*,\lambda})}{\partial{P}_{U|X}}=\kappa_x-\phi_{x,u}$ for all $(u,x)\in\mathcal{U}\times\mathcal{X}$,
	\item $\sum_u{P}_{U|X}^{*,\lambda}=1$ for all $x\in\mathcal{X}$,
	\item ${P}_{U|X}^{*,\lambda}\geq0$ for all $(u,x)\in\mathcal{U}\times\mathcal{X}$,
	\item $\phi_{x,u}\geq0$ for all $(u,x)\in\mathcal{U}\times\mathcal{X}$,
	\item $\phi_{x,u}{P}_{U|X}^{*,\lambda}=0$ for all $(u,x)\in\mathcal{U}\times\mathcal{X}$.
\end{enumerate}
From conditions 1) and 4), we obtain for all $(u,x)\in\mathcal{U}\times\mathcal{X}$:
\begin{equation}
\kappa_x\geq \frac{\partial f(\lambda,{P}_{U|X}^{*,\lambda})}{\partial{P}_{U|X}}.	
\end{equation}
From condition 5), we observe that equality is achieved for all  $(u,x)\in\mathcal{U}\times\mathcal{X}$ such that ${P}_{U|X}^{*,\lambda}>0$. From (\ref{derivadaf}) we have:
\begin{align}
\frac{\partial f(\lambda,{P}_{U|X}^{*,\lambda})}{\partial{P}_{U|X}}&=(2\lambda-1)\sum_z{P}_{XZ}\log\left(\frac{{Q}^{*}_{X|ZU}}{{P}_{X|Z}}\right)\nonumber\\
&\quad-\lambda\sum_{y,z}{P}_{XYZ}\log\left(\frac{{P}_{U|X}^{*,\lambda}}{{Q}^{*}_{U|YZ}}\right).
\end{align}
Combining these two last equations and summing over all $x\in\mathcal{X}$, we have:
\begin{align}
\sum_x \kappa_x&\geq (2\lambda-1)\sum_{x,z}{P}_{XZ}\log\left(\frac{{Q}^{*}_{X|ZU}}{{P}_{X|Z}}\right)\nonumber\\
&\quad-\lambda\sum_{x,y,z}{P}_{XYZ}\log\left(\frac{{P}_{U|X}^{*,\lambda}}{{Q}^{*}_{U|YZ}}\right)\\
&=\alpha^{*}(\lambda,u).
\end{align}
In a similar manner, from conditions 1) and 5) and using Eq. \eqref{igualdadkkt}, we can write:
\begin{align}
V_\lambda=f(\lambda,{P}_{U|X}^{*,\lambda})&=\sum_{x,u}{P}_{U|X}^{*,\lambda}\frac{\partial f(\lambda,{P}_{U|X}^{*,\lambda})}{\partial{P}_{U|X}}\\
&=\sum_{x,u}{P}_{U|X}^{*,\lambda}\left(\kappa_x-\phi_{x,u}\right)\\
&=\sum_x\kappa_x.
\end{align}
It is straightforward to check that $V_\lambda\geq \alpha^{*}(\lambda,u)$ for all $u\in\mathcal{U}$. Finally, from condition 5) and by similar arguments, it is easy to see that $V_\lambda=\alpha^{*}(\lambda,u)$ $\forall$ $(u,x)\in\mathcal{U}\times\mathcal{X}$ s.t. ${P}_{U|X}^{*,\lambda}>0$.

\section*{Appendix D: Convergence rate}~\label{app:convergence-rate}
The speed of convergence can be obtained easily from the proof of Theorem \ref{convergencia}.
\begin{lemma}
	Consider $\lambda\in(0.5,1]$ and ${P}_{U|X}^{(0)}\in G_{\delta,\lambda}$ for a given value of $\delta$. If the optimal solution $P_{U|X}^{*,\lambda}$ lies in $H_{\delta,\lambda}({P}_{U|X}^{(0)})$ and the function $f(\lambda,{P}_{U|X})$ is concave in $H_{\delta,\lambda}({P}_{U|X}^{(0)})$ and ${P}_{U|X}^{(0)}$ is such that $|\text{supp}({P}_{U|X}^{(0)})|=|\mathcal{U}|$, we have:
	\begin{equation}
	V_\lambda-F_\lambda^{(N)}\leq\frac{\lambda}{N}\cdot\mathbb{E}_{X}[\mathcal{D}(P_{U|X}^{*,\lambda}\|{P}_{U|X}^{(0)})].
	\end{equation}
\end{lemma}
\begin{IEEEproof}
	$F_\lambda^{(n)}$ is monotonically non-decreasing and is bounded by $V_\lambda$. Then,
	\begin{equation}
	\sum_{n=0}^{N-1}V_\lambda-F_\lambda^{(n+1)}\geq N(V_\lambda-F_\lambda^{(N)}).
	\end{equation}
	The LHS term can be bounded using \eqref{ec:sumbound}:
	\begin{equation}
	\lambda\mathbb{E}_X\left[\mathcal{D}(P_{U|X}^{*,\lambda}\|{P}_{U|X}^{(0)})\right]\geq N(V_\lambda-F_\lambda^{(N)}).
	\end{equation}
	Finally,
	\begin{equation}
	V_\lambda-F_\lambda^{(N)}\leq\frac{\lambda}{N}\cdot\mathbb{E}_X[\mathcal{D}(P_{U|X}^{*,\lambda}\|{P}_{U|X}^{(0)})].
	\end{equation}
\end{IEEEproof}
We can see that the approximation error $V_\lambda-F_\lambda^{(N)}$ is inversely proportional to the number of iterations and the algorithm has a rate of convergence of at least order $1/N$.

\bibliographystyle{IEEEtran}
\bibliography{matias.bib}

\end{document}